\documentclass[10pt, twocolumn, letterpaper, twosided]{article}
\usepackage[top=1in, bottom=1in, left=1in, right=1in]{geometry} 

\usepackage{fancyhdr}
\usepackage{amssymb, amsmath, amsthm}
\usepackage{graphicx}
\usepackage{color}
\usepackage{xspace}
\usepackage[tight]{subfigure}
\usepackage[ruled,vlined]{algorithm2e}
\usepackage{array}
\usepackage{times}
\usepackage{url}

\newcommand{\lessdenselist}{
  \itemsep -3pt 
}

\newcommand{\denselistbib}{
  \itemsep 0pt\topsep-4pt\partopsep-5pt
}

\newcommand{\tightsection}[1]{\vspace{-4mm}
  \section{#1}\vspace{-1mm}}

\newcommand{\tightsubsection}[1]{
  \vspace{-2mm}
  \subsection{#1}
  \vspace{-1mm}
}

\graphicspath{{figures/}{experiments/}}

\numberwithin{equation}{section}

\newtheorem{dfn}{Definition}[section]

\newtheorem{thm}{Theorem}[section]

\newtheorem{ex}[thm]{Example}
\newtheorem*{exa}{Example}

\newcommand{\supscriptbf}[1]{\large\ensuremath{^{\textrm{\textbf{#1}}}}}

\newcolumntype{C}[1]{>{\centering\arraybackslash}p{#1}}

\newcommand{\notestojoe}[1]{}

 \newcommand{\joey}[1]{}
 \newcommand{\yucheng}[1]{}
 \newcommand{\carlos}[1]{}
 \newcommand{\aapo}[1]{}
 \newcommand{\db}[1]{}

\newcommand{\term}[1]{\textbf{#1}}

\newcommand{\tableref}[1]{Table~\ref{#1}}
\newcommand{\figref}[1]{Fig.~\ref{#1}}
\newcommand{\eqnref}[1]{Eq.~(\ref{#1})}
\newcommand{\secref}[1]{Sec.~\ref{#1}}

\newcommand{\algref}[1]{Alg.~\ref{#1}}

\newcommand{\alglineref}[1]{Line~\ref{#1}}

\newcommand{\exampleref}[1]{Ex.~\ref{#1}}

\newcommand{\abs}[1]{\left|#1\right|}

\newcommand{\set}[1]{\left\{#1\right\}}

\newcommand{\union}{\cup}

\newcommand{\size}[1]{\left| #1 \right|}

\newcommand{\BigO}[1]{O \left( #1 \right)}

\newcommand{\neighbors}[1]{\mathbf{N}[ #1 ]}

\newcommand{\alldata}{D}
\newcommand{\vertexdata}[1]{D_{#1}}

\newcommand{\edgedata}[2]{D_{#1 \rightarrow #2}}
\newcommand{\biedgedata}[2]{D_{#1 \leftrightarrow #2}}

\newcommand{\accum}{\text{acc}}
\newcommand{\fold}{\textbf{Fold}}
\newcommand{\merge}{\textbf{Merge}}
\newcommand{\finalize}{\textbf{Finalize}}
\newcommand{\syncinterval}{\tau}
\newcommand{\updatefunction}{\textbf{Update}}
\newcommand{\scope}[1]{\mathcal{S}_{#1}}

\newcommand{\PageRank}{\textbf{R}}

\SetKwInOut{Input}{Input}
\SetKwInOut{Output}{Output}
\SetKwFunction{Residual}{Residual}
\SetKwFor{ParForAll}{forall}{do in parallel}{}
\SetKwFunction{Reverse}{ReverseOrder}
\SetKwBlock{Sync}{Synchronized}{end}
\SetKwFunction{Map}{Map}
\SetKwFunction{Reduce}{Reduce}
\SetKwFunction{Push}{Push}
\SetKwFunction{InitializeQueue}{InitializeQueue}
\SetKwFunction{TopResid}{TopResid}
\SetKwFunction{Top}{Top}
\SetKwFunction{Pop}{Pop}
\SetKwFunction{RemoveNext}{RemoveNext}
\SetKwFunction{Enqueue}{Enqueue}
\SetKwFunction{Dequeue}{Dequeue}
\SetKwFunction{UpdatePriority}{UpdatePriority}
\SetKwFunction{SetPriority}{SetPriority}
\SetKwFor{Lock}{lock}{}{release}
\SetKwFunction{Collect}{Collect}
\SetKwFunction{Initialize}{Initialize}
\SetKwFunction{TokenRing}{TokenRing}
\SetKwFunction{Promote}{Promote}
\SetKwFunction{Value}{Value}

\title{GraphLab: A Distributed Framework for\\ Machine Learning in the
  Cloud} 
\author{Yucheng Low, Joseph Gonzalez, Aapo Kyrola, Danny Bickson, Carlos Guestrin}
\date{}

\begin{document}
\maketitle
\thispagestyle{fancy}
\begin{abstract}
  Machine Learning (ML) techniques are indispensable
  in a wide range of fields.
  Unfortunately, the exponential increase of dataset sizes are rapidly
  extending the runtime of sequential algorithms and
  threatening to slow future progress in ML.  With the promise of
  affordable large-scale parallel computing, Cloud systems offer a
  viable platform
  to resolve the computational challenges in ML.  However, designing
  and implementing \emph{efficient}, \emph{provably correct}
  distributed ML algorithms is often prohibitively challenging.  To
  enable ML researchers to easily and efficiently use parallel
  systems, we introduced the GraphLab abstraction which is designed to
  represent the computational patterns in ML algorithms while permitting
  efficient parallel and distributed implementations.

  In this paper we provide a formal description of the GraphLab
  parallel abstraction and present an efficient distributed
  implementation.  We conduct a comprehensive evaluation of GraphLab
  on three state-of-the-art ML algorithms
  using real large-scale data and a 64 node EC2 cluster of 512 processors.  We
  find that GraphLab achieves orders of magnitude performance gains
  over Hadoop while performing comparably or superior to hand-tuned
  MPI implementations.
\end{abstract}

\tightsection{Introduction}

With the exponential growth in Machine Learning (ML) datasets sizes
and increasing sophistication of ML techniques, there is a growing
need for systems that can execute ML algorithms efficiently in
parallel on large clusters.  Unfortunately, based on our comprehensive
survey, we find that existing popular high level parallel
abstractions, such as MapReduce \cite{dean04, cheng06} and Dryad
\cite{isard2007dryad}, do not efficiently fit many ML applications.
Alternatively, designing, implementing, and debugging ML algorithms on
low level frameworks such as OpenMP \cite{dagum1998openmp} or MPI
\cite{gropp1996high} can be excessively challenging, requiring the
user to address complex issues like race conditions, deadlocks, and
message passing in addition to the already challenging
mathematical code and complex data models common in ML research.

In this paper we describe the culmination of two years of research in
collaboration with ML, Parallel Computing, and Distributed Systems
experts.
By focusing on Machine Learning we designed GraphLab, a
domain specific parallel abstraction \cite{asanovic2006landscape} that
fits the needs of the ML community, without sacrificing
computational efficiency or requiring ML researchers to redesign their
algorithms.
In \cite{uaigraphlab_anon} we first introduced the GraphLab multi-core
API to the ML community.

In this paper we build upon our earlier work by refining the GraphLab
abstraction and extending the GraphLab API to the distributed setting.
We provide the first formal presentation of the streamlined GraphLab
abstraction and describe how the abstraction enabled us to
construct a highly optimized C++ API for the distributed
setting.  We conduct a comprehensive performance analysis on the
Amazon Elastic Cloud (EC2) cloud computing service. We show that
applications created using GraphLab outperform equivalent
Hadoop/MapReduce\cite{dean04} implementations by 20-60x and match the
performance of carefully constructed and fine tuned MPI
implementations.  Our main contributions are the following:

\begin{itemize}
  \lessdenselist

\item Asurvey of common properties of Machine Learning algorithms and
  the limitations of existing parallel abstractions.
  (\secref{sec:needforgraphlab})

\item A formal presentation of the GraphLab abstraction and how it
  naturally represents ML algorithms. (\secref{sec:graphlababstraction})

\item Two efficient distributed implementations of the GraphLab
  abstraction (\secref{sec:impldetails}):
  \begin{list}{$\circ$}{
      \setlength{\leftmargin}{1em} 
      \setlength{\itemsep}{-1pt}
      \setlength{\topsep}{-2pt} 
    }
  \item \textbf{Chromatic Engine:} uses graph coloring to achieve
    efficient sequentially consistent execution for static schedules.
  \item \textbf{Locking Engine:} uses distributed locking and latency
    hiding to achieve sequential consistency while supporting
    prioritized execution.
  \end{list}
\item Implementations of three state-of-the-art machine learning
  algorithms using the GraphLab abstraction. (\secref{sec:applications})
\item An extensive evaluation of GraphLab using a 512 processor (64 node) EC2
  cluster, including comparisons to Hadoop and MPI
  implementations. (\secref{sec:experiments})
\end{itemize}

\tightsection{A Need for GraphLab in ML}
\label{sec:needforgraphlab}

\begin{table*}[t]
\begin{small}
\begin{center}
\begin{tabular}{| l || C{0.7in} | C{0.5in} | C{0.5in} | C{0.6in} | C{0.6in} | C{0.7in}  | C{0.6in}| } 
\hline   & \textbf{Computation Model} & \textbf{Sparse Depend.} & \textbf{Async. Comp.} & \textbf{Iterative} & \textbf{Prioritized Ordering} &  \textbf{Sequentially Consistent \supscriptbf{a}} & \textbf{Distributed} \\
\hline
\hline MPI\cite{gropp1996high} & Messaging & Yes & Yes & Yes & N/A  \supscriptbf{b} & N/A  \supscriptbf{b}& Yes \\
\hline MapReduce\cite{dean04} & Par. data-flow& No & No & extensions\supscriptbf{c} & N/A  & N/A  & Yes \\
\hline Dryad\cite{isard2007dryad} & Par. data-flow &Yes & No & extensions\supscriptbf{d} &  N/A   & N/A  & Yes \\
\hline Pregel\cite{pregel}/BPGL\cite{gregor2005parallel} &  GraphBSP\cite{valiant1990bridging} & Yes & No & Yes & N/A  & N/A & Yes  \\
\hline Piccolo\cite{power2010piccolo} & Distr. map\supscriptbf{f} & N/A \supscriptbf{f} & Yes & Yes & No & accumulators & Yes \\
\hline Pearce et.al.\cite{pearce2010multithreaded} & Graph Visitor & Yes & Yes & Yes & Yes & No & No \\
\hline\hline \textbf{GraphLab} & \textbf{GraphLab} & Yes & Yes & Yes & Yes\supscriptbf{e} & Yes & Yes\\
\hline
\end{tabular}
\vspace{-2mm}
\caption{ \footnotesize \textbf{Comparison chart of parallel
    abstractions: }  Detailed comparison against each
  of the abstractions are in the text
  (\secref{sec:needforgraphlab}, \secref{sec:relatedwork}).  \textbf{(a)} Here we refer to
  Sequential Consistency with respect to asynchronous computation.
  See \secref{sec:needforgraphlab} for details. This property is therefore
  relevant only for abstractions which support asynchronous
  computation.  \textbf{(b)} MPI-2 does not define a data model and is
  a lower level abstraction than others listed.  \textbf{(c)}
  Iterative extension for MapReduce are proposed
  \cite{mapreduceonline, hindman2009common, zaharia2010spark}.
  \textbf{(d)} \cite{hindman2009common} proposes an iterative
  extension for Dryad.  \textbf{(e)} The GraphLab abstraction allows for flexible scheduling mechanisms (our implementation provides FIFO and priority
  ordering).  \textbf{(f)} Piccolo computes using user-defined kernels
  with random access to a distributed key-value store. It does not model data dependencies.
  }
\label{table:abstrcomptable}
\end{center}
\end{small}
\vspace{-6mm}
\end{table*}

The GraphLab abstraction is the product of several years of research
in designing and implementing systems for statistical inference in
probabilistic graphical models. Early in our work
\cite{gonzalez09_anon}, we discovered that the high-level parallel
abstractions popular in the ML community such as MapReduce
\cite{dean04, cheng06} and parallel BLAS \cite{choi1996proposal}
libraries are unable to express statistical inference algorithms
efficiently.
Our work revealed that an efficient algorithm for graphical model
inference should explicitly address the \emph{sparse dependencies}
between random variables and adapt to the input data and model
parameters.

Guided by this intuition we spent over a year designing and
implementing various machine learning algorithms on top of low-level
threading primitives and distributed communication frameworks such as
OpenMP \cite{dagum1998openmp}, CILK++ \cite{leiserson2009cilkpp++} and
MPI \cite{gropp1996high}.  Through this process, we discovered the
following set of core algorithmic patterns that are common to a wide
range of machine learning techniques. Following, we detail our findings and motivate why a new framework is needed
(see \tableref{table:abstrcomptable}).



\textbf{Sparse Computational Dependencies:} Many ML algorithms can be
factorized into local \term{dependent} computations which examine and
modify only a small sub-region of the entire program state.  For
example, the conditional distribution of each random variable in a
large statistical model typically only depends on a small subset of
the remaining variables in the model.
This computational sparsity in machine learning arises naturally from
the statistical need to reduce model
complexity.

Parallel abstractions like MapReduce \cite{dean04} require algorithms
to be transformed into an embarrassingly parallel form where
computation is \term{independent}.  Unfortunately, transforming ML
algorithms with computational \emph{dependencies} into the
embarrassingly parallel form needed for these abstractions is often
complicated and can introduce substantial algorithmic inefficiency
\cite{aistats}.  Alternatively, data flow abstractions like Dryad
\cite{isard2007dryad}, permit directed acyclic dependencies, but
struggle to represent cyclic dependencies common to iterative ML
algorithms.  Finally, graph-based messaging abstractions like Pregel
\cite{pregel} provide a more natural representation of computational
dependencies but require users to explicitly manage communication
between computation units.

 


\textbf{Asynchronous Iterative Computation:} From simulating complex
statistical models, to optimizing
parameters, 
many important machine learning algorithms iterate over local
computation kernels. Furthermore, many iterative machine learning
algorithms benefit from \cite{bertsekas, Siapas96, gonzalez09_anon}
and in some cases require \cite{gonzalez11} asynchronous computation.
Unlike \term{synchronous} computation, in which all kernels are
computed simultaneously (in parallel) using the previous values for
dependent parameters, \term{asynchronous} computation requires that
the local computation kernels use the most recently available values.

Abstractions based on bulk data processing, such as MapReduce
\cite{dean04} and Dryad \cite{isard2007dryad} were not designed for
iterative computation.  While recent projects like MapReduce Online
\cite{mapreduceonline}, Spark \cite{zaharia2010spark}, Twister
\cite{ekanayake2010twister}, and Nexus \cite{hindman2009common}
extend MapReduce to the iterative setting, they do not
support asynchronous computation.  Similarly, parallel graph based
abstractions like Pregel \cite{pregel} and BPGL
\cite{gregor2005parallel} adopt the Bulk Synchronous Parallel (BSP)
model \cite{valiant1990bridging} and do not naturally express
asynchronous computation.


%
%




\textbf{Sequential Consistency:} By ensuring that all parallel
executions have an equivalent sequential execution, sequential
consistency eliminates many challenges associated with designing,
implementing, and testing parallel ML algorithms. In addition, 
many algorithms converge faster if sequential consistency is ensured, 
and some even require it for correctness. 

However, this view is not shared by all in the ML community. 
Recently,  \cite{besteffort09, Chafi11} advocate soft-optimization
techniques (e.g., allowing computation to intentionally race), but we
argue that such techniques do not apply broadly in ML.  Even for
the algorithms evaluated in \cite{besteffort09, Chafi11}, the
conditions under which the soft-optimization techniques work are not
well understood and may fail in unexpected ways on different datasets.

\begin{figure}[b]
  \begin{center}
          \includegraphics[width=0.25\textwidth]{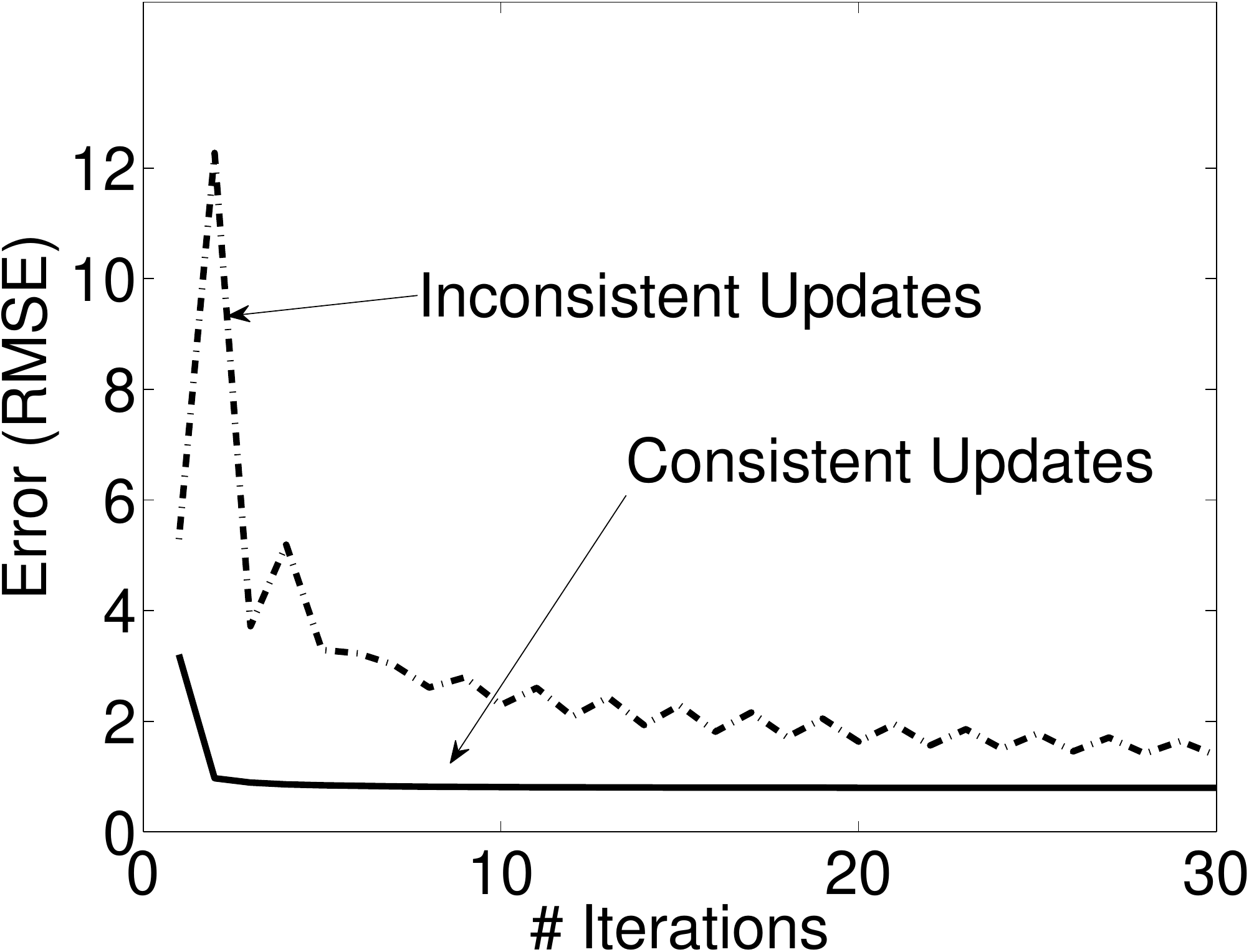}
  \vspace{-2mm}
          \caption{\footnotesize 
            Convergence plot of Alternating Least Squares (\secref{sec:applications})
            comparing prediction error when running sequentially
            consistent asynchronous iterations vs inconsistent
            asynchronous iterations over a five node distributed
            cluster. Consistent iterations converge rapidly to a lower
            error while inconsistent iterations oscillate and converge
            slowly.}
            \label{fig:synchronouspmf}
  \end{center}
  \vspace{-2mm}
\end{figure}

Indeed, for some machine learning algorithms sequential consistency is
strictly required.  For instance, Gibbs sampling \cite{Geman84}, a
popular inference algorithm, requires sequential consistency for statistical
correctness, while many other optimization procedures require
sequential consistency to converge (\figref{fig:synchronouspmf}
demonstrates that the prediction error rate of one of our example problems is
dramatically better when computation is properly asynchronous.).
Finally, as \cite{Siapas96} demonstrates, the lack of sequential
consistency can dramatically increase the time to convergence for
stochastic optimization procedures.

By designing an abstraction which enforces sequentially consistent
computation, we eliminate much of the complexity introduced by parallelism,
allowing the ML expert to focus on algorithm design and correctness
of numerical computations. Debugging mathematical code in a parallel
program which has random errors caused by non-deterministic ordering 
of concurrent computation is particularly unproductive.

The discussion of sequential consistency is relevant only to frameworks which support
asynchronous computation. Piccolo \cite{power2010piccolo} provides 
a limited amount of consistency by combining simultaneous writes using
accumulation functions. However, this only protects against
single write races, but does not ensure sequential consistency in general.
The parallel asynchronous graph traversal abstraction by Pearce et. al. \cite{pearce2010multithreaded} 
does not support any form of consistency, and thus is not suitable for a large
class of ML algorithms.

\textbf{Prioritized Ordering:} In many ML algorithms, iterative
computation converges asymmetrically.  For example, in parameter
optimization, often a large number of parameters will quickly converge
after only a few iterations, while the remaining parameters will
converge slowly over many iterations \cite{elidan06, EfrHasJohTib04}.
If we update all parameters equally often, we could waste substantial
computation recomputing parameters that have effectively converged.
Conversely, by focusing early computation on more challenging
parameters first, we can potentially reduce computation.


Adaptive prioritization can be used to focus iterative computation
where it is needed.  The only existing framework to support this is 
the parallel graph framework by Pearce et. al. 
\cite{pearce2010multithreaded}. The framework is based on the 
visitor-pattern and prioritizes the ordering of visits to
vertices.  
GraphLab however, allows the user to define \emph{arbitrary ordering} of 
computation, and our implementation supports
efficient FIFO and priority-based scheduling.  

%

%


\textbf{Rapid Development:} 
Machine learning is a rapidly evolving field with new algorithms and
data-sets appearing weekly.  In many cases these algorithms are not
yet well characterized and both the computational and statistical
properties are under active investigation.
Large-scale parallel machine learning systems must be able to adapt
quickly to changes in the data and models in order to facilitate rapid
prototyping, experimental analysis, and model tuning.  To achieve
these goals, an effective high-level parallel abstraction must hide
the challenges of parallel algorithm design, including race conditions,
deadlock, state-partitioning, and communication.

\tightsection{The GraphLab Abstraction}
\label{sec:graphlababstraction}
Using the ideas from the previous section, we extracted a single
coherent computational pattern: \emph{asynchronous parallel
  computation on graphs} with a \emph{sequential} model of
computation.  This pattern is both sufficiently expressive to
encode a wide range of ML algorithms, and sufficiently restrictive
to enable efficient parallel implementations.


The GraphLab abstraction consists of three main parts, the data graph,
the update function, and the sync operation.  The data graph
(\secref{sec:data_graph}) represents user modifiable program state,
and both stores the mutable user-defined data and encodes the sparse
computational dependencies. The update functions
(\secref{sec:update_functions}) represent the factorized user
computation and operate on the data graph by transforming data in
small overlapping contexts called scopes.  Finally, the sync operation
(\secref{sec:sync_operation}) is used to maintain global aggregate
statistics of the data graph.

We now present the GraphLab abstraction in greater detail.  To make
these ideas more concrete, we will use the
PageRank algorithm \cite{pagerank} as a running example.
 While PageRank is not a common machine learning algorithm, it is
 easy to understand and shares many properties common to machine
 learning algorithms.
\begin{ex}[PageRank]
  \label{ex:pagerank_intro}
  The PageRank algorithm recursively defines the rank of a webpage
  $v$:
  \begin{equation}
    \PageRank( v ) = 
    \frac{\alpha}{n}
    + (1-\alpha) \sum_{\text{$u$ links to $v$}}w_{u,v}  \times \PageRank(u) 
    \label{eqn:pagerank}
  \end{equation}
  in terms of the ranks of those pages that link to $v$ and the weight
  $w$ of the link as well as some probability $\alpha$ of randomly
  jumping to that page. The PageRank algorithm, simply iterates
  \eqnref{eqn:pagerank} until the individual PageRank values converge
  (i.e., change by less than some small $\epsilon$).
\end{ex}

\tightsubsection{Data Graph}
\label{sec:data_graph}


The GraphLab abstraction stores the program state as an undirected
graph called the \term{data graph}.  The data graph $G=(V, E,
\alldata)$ is a container which manages the user defined data $D$.
Here we use the term ``data" broadly to refer to model parameters,
algorithmic state, and even statistical data.  The user can associate
arbitrary data with each vertex $\set{\vertexdata{v} : v \in V}$ and
edge $\set{\biedgedata{u}{v} : \set{u,v} \in E}$ in the graph.  Since
some machine learning applications require directed edge data (e.g.,
weights on directed links in a web-graph) we provide the ability to
store and retrieve data associated with directed edges.  While the
graph data is mutable, the graph structure is
\emph{static}\footnote{Although we find that fixed structures are
  sufficient for most ML algorithms, we are currently exploring the
  use of dynamic graphs.} and cannot be changed during execution.


\begin{exa}[PageRank: \exampleref{ex:pagerank_intro}]
  The data graph for PageRank is directly
  obtained from the web graph, where each vertex corresponds to a web
  page and each edge represents a link.  The vertex data
  $\vertexdata{v}$ stores $\PageRank(v)$, the current estimate of the
  PageRank, and the edge data $\edgedata{u}{v}$ stores $w_{u,v}$, the
  directed weight of the link. 
\end{exa}

The data graph is convenient for representing the state of a wide
range of machine learning algorithms.  For example, many statistical
models are efficiently represented by undirected graphs
\cite{friedmankoller} called Markov Random Fields (MRF).  The data
graph is derived directly from the MRF, with each vertex representing
a random variable. In this case the vertex data and edge data
may store the local parameters that we are interested in learning.

\tightsubsection{Update Functions}
\label{sec:update_functions}
%

\begin{figure}[t]
	\centering
    \includegraphics[width=.30\textwidth,clip]{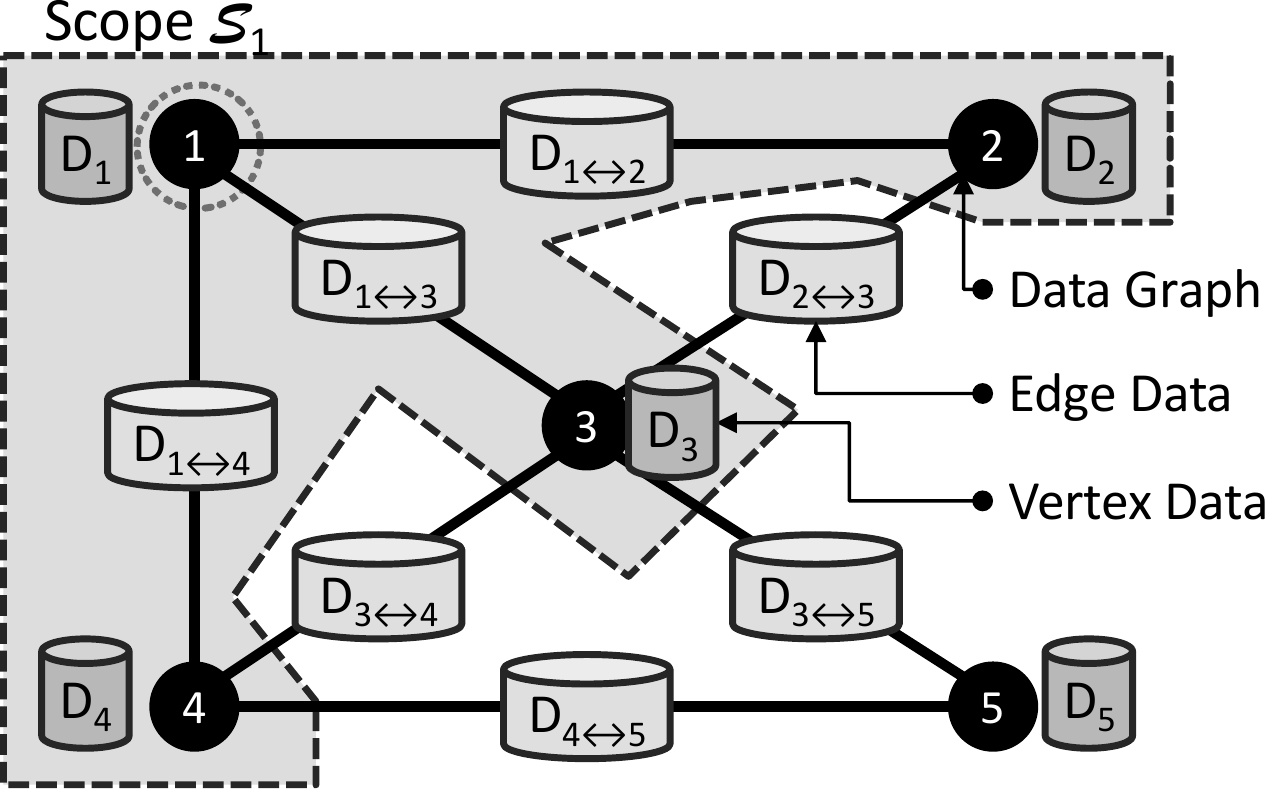}
    \label{fig:scope}
  \vspace{-2mm}
  \caption{\footnotesize In this figure we illustrate the GraphLab
    \term{data graph} as well as the scope $\scope{1}$ of vertex $1$.
    Each of the gray cylinders represent a block of user defined data
    and is associated with a vertex or edge.  The \term{scope} of
    vertex $1$ is illustrated by the region containing vertices
    $\set{1,2,3,4}$.  An \term{update function} applied to vertex $1$
    is able to read and modify all the data in $\scope{1}$ 
    (vertex data $\vertexdata{1}$, $\vertexdata{2}$, $\vertexdata{3}$,
    and $\vertexdata{4}$ and edge data $\biedgedata{1}{2}$,
    $\biedgedata{1}{3}$, and $\biedgedata{1}{4}$).  }
\end{figure}

Computation is encoded in the GraphLab abstraction via user defined
update functions.  An \term{update function} is a stateless procedure
which modifies the data within the \term{scope} of a vertex and
schedules the future execution of other update functions.  The scope
of vertex $v$ (denoted by $\scope{v}$) is the data stored in $v$, 
as well as the data stored in all adjacent vertices and edges
as shown in \figref{fig:scope}.



A GraphLab update function 
takes as an input a vertex $v$ and its scope $\scope{v}$ and returns
the new version of the scope as well as a set of tasks $\mathcal{T}$
which encodes future task executions.
\[
 \updatefunction: (v, \scope{v}) \rightarrow 
  \left( \scope{v}, \mathcal{T} \right)
\]
After executing an update function the modified scope data in $\scope{v}$ is
written back to the data graph.  Each \term{task} in the set of tasks
$\mathcal{T}$, is a tuple $(f, v)$ consisting of an update function
$f$ and a vertex $v$.  All returned task $\mathcal{T}$ are executed
\emph{eventually} by running $f(v, \scope{v})$ following the execution
semantics described in \secref{sec:execution_model}.


Rather than adopting a message passing or data flow model as in
\cite{pregel, isard2007dryad}, GraphLab allows the user defined update
functions complete freedom to read and modify any of the data on
adjacent vertices and edges.  This simplifies user code and eliminates
the need for the users to reason about the movement of data.
By controlling what tasks are added to the task set, GraphLab update
functions can efficiently express adaptive computation.  For example,
an update function may choose to reschedule its neighbors only when it
has made a substantial change to its local data.

The update function mechanism allows for \emph{asynchronous
computation} on the \emph{sparse dependencies} defined by the 
data graph. Since the data graph permits the expression of 
general cyclic  dependencies,
\emph{iterative computation} can be represented easily.

Many algorithms in machine learning can be expressed as
simple update functions.  For example, probabilistic inference
algorithms like Gibbs sampling \cite{Geman84}, belief propagation
\cite{pearl88}, expectation propagation \cite{Minka01} and mean field
variational inference \cite{Xing03} can all be expressed using update
functions which read the current assignments to the parameter
estimates on neighboring vertices and edges and then apply
sampling or optimization techniques to update parameters on
the local vertex.

\begin{exa}[PageRank: \exampleref{ex:pagerank_intro}]
  The update function for PageRank (defined in
  \algref{alg:pagerank_updatefunction}) computes a weighted sum of the
  current ranks of neighboring vertices and assigns it as the rank of
  the current vertex.  The algorithm is adaptive: neighbors are listed
  for update only if the value of current vertex changes more than a
  predefined threshold.  


\end{exa}

\begin{algorithm}[t]
  \label{alg:pagerank_updatefunction}
  \caption{PageRank update function}
  \DontPrintSemicolon
  \SetKwFunction{Reschedule}{Reschedule}
  \begin{small}

    \KwIn{Vertex data $\PageRank(v)$ from $\scope{v}$}

    \KwIn{Edge data $\set{w_{u,v} : u \in \neighbors{v}}$ from
      $\scope{v}$}

    \KwIn{Neighbor vertex data $\set{\PageRank(u) : u \in
        \neighbors{v}}$ from $\scope{v}$}

    $\PageRank_{\text{old}}(v) \leftarrow \PageRank(v) $ 
    \tcp*[f]{Save old PageRank} $\PageRank(v) \leftarrow
    \alpha/n$\;

    \ForEach(\tcp*[f]{Loop over neighbors}){$u \in \neighbors{v}$ }{  
      $\PageRank(v) \leftarrow \PageRank(v) + (1-\alpha) * w_{u,v} *
      \PageRank(u)$\;
    }
    \tcp{If the PageRank changes sufficiently}
    \If{$\abs{\PageRank(v) - \PageRank_{\text{old}}(v)}  >\epsilon $} {
      \tcp{Schedule neighbors to be updated}
      \Return{$\set{(\texttt{PageRankFun}, u) : u \in \neighbors{v}}$}
    }
    \KwOut{Modified scope $\scope{v}$ with new $\PageRank(v)$}
  \end{small}
\end{algorithm}




\tightsubsection{Sync Operation}
\label{sec:sync_operation}

In many ML algorithms it is necessary to maintain global
statistics describing data stored in the data graph.  For example, many
statistical inference algorithms require  tracking of global
convergence estimators.  Alternatively, parameter estimation
algorithms often compute global averages or even gradients to tune
model parameters.  To address these situations, the GraphLab
abstraction expresses global computation through the \term{sync
  operation}, which aggregates data across all vertices in the graph
in a manner analogous to MapReduce.  The results of the sync operation
are stored globally and may be accessed by all update functions.
Because GraphLab is designed to express iterative computation, the
sync operation runs repeated at fixed user determined intervals to
ensure that the global estimators remain fresh.



The sync operation is defined as a tuple $(\texttt{Key}, \fold,
\merge, \finalize, \accum(0), \syncinterval)$ consisting of a unique
key, three user defined functions, an initial accumulator value, and
an integer defining the interval between sync operations.  The sync
operation uses the $\fold$ and $\merge$ functions to perform a
\emph{Global Synchronous Reduce} where $\fold$ aggregates
vertex data and $\merge$ combines intermediate $\fold$ results.  The
$\finalize$ function performs a transformation on the final value and
stores the result.  The \texttt{Key} can then be used by
update functions to access the most recent result of the sync
operation.  The sync operation runs periodically, approximately every
$\syncinterval$ update function calls\footnote{The resolution
  of the synchronization interval is left up to the implementation
  since in some architectures a precise synchronization interval may
  be difficult to maintain.}.

\begin{exa}[PageRank: \exampleref{ex:pagerank_intro}]  
  We can compute the second most popular page on the web
  by defining the following sync operation:
  \begin{align*}
    \fold: & \text{fld}(\accum, v, \vertexdata{v}) :=
    \text{TopTwo}(\accum \union \PageRank(v)) \\
    \merge: & \text{mrg}(\accum, \accum^\prime) :=
    \text{TopTwo}(\accum \union  \accum^\prime) \\
    \finalize: & \text{fin}(\accum) := \accum[2]
  \end{align*}
  Where the accumulator taking on the initial value as the 
  empty array $\accum(0) = \emptyset$.
  The function ``$\text{TopTwo}(X)$" returns the two pages with the highest pagerank in the set $X$.
  After the global reduction, the $\accum$ array will contain the top two pages and 
  $\accum[2]$ in $\finalize$ extracts the second entry.
  We may want to update the global estimate every $\syncinterval =
  \size{V}$ vertex updates.
\end{exa}



\tightsubsection{The GraphLab Execution Model}
\label{sec:execution_model}
The GraphLab execution model, presented in
\algref{alg:execution_model}, follows a simple single loop semantics.
The input to the GraphLab abstraction consists of the data graph $G =
(V, E, \alldata)$, an update function $\updatefunction$, an initial
set of tasks $\mathcal{T}$ to update, and any sync operations. While
there are tasks remaining in $\mathcal{T}$, the algorithm removes
(\alglineref{ln:pop}) and executes (\alglineref{ln:update}) tasks,
adding any new tasks back into $\mathcal{T}$ (\alglineref{ln:push}).
The appropriate sync operations are executed whenever necessary.  Upon
completion, the resulting data graph and synced values are returned to
the user.

\begin{algorithm}[t]
  \label{alg:execution_model}
  \caption{GraphLab Execution Model}
  \DontPrintSemicolon
  \KwIn{Data Graph $G=(V, E, \alldata)$ }
  \KwIn{Initial task set $\mathcal{T}= \set{(f,v_1), (g, v_2), ...}$}

  \KwIn{Initial set of syncs:  $\left(\text{Name}, \fold, \merge, \finalize,
      \accum(0), \syncinterval \right)$}

  \While{$\mathcal{T}$ is not Empty }{
    \lnl{ln:pop} 
    $(f,v) \leftarrow $ \RemoveNext{$\mathcal{T}$} \;
    \lnl{ln:update} 
    $(\mathcal{T}^\prime, \scope{v}) \leftarrow f(v, \scope{v})$ \;
    \lnl{ln:push} 
    $\mathcal{T} \leftarrow \mathcal{T} \union \mathcal{T}^\prime$ \;
    Run all Sync operations which are ready \;
  }
  \KwOut{Modified Data Graph $G=(V, E, D^\prime)$}
  \KwOut{Result of Sync operations}
\end{algorithm}

The exact behavior of $\RemoveNext(\mathcal{T})$ (\alglineref{ln:pop})
is up to the implementation of the GraphLab abstraction.  The only
guarantee the GraphLab abstraction provides is that $\RemoveNext$
removes and returns an update task in $\mathcal{T}$.  The flexibility
in the order in which $\RemoveNext$ removes tasks from $\mathcal{T}$
provides the opportunity to balance features with performance constraints.
For example, by restricting task execution to a fixed
order, it is possible to optimize memory layout.  Conversely, by
supporting \emph{prioritized ordering} it is possible to implement more
advanced ML algorithms at the expense run-time overhead.  
In our implementation (see \secref{sec:impldetails}) we support 
fixed execution ordering (Chromatic Engine) as well as FIFO 
and prioritized ordering (Locking Engine).



The GraphLab abstraction presents a rich \emph{sequential model} that
is automatically translated into a \emph{parallel execution} 
by allowing multiple processors to remove and execute update tasks
simultaneously. 
To retain the same \emph{sequential} execution semantics we must ensure that
overlapping computation is not run simultaneously.  However, the
extent to which computation can \emph{safely} overlap depends on the
user defined update function.  In the next section we introduce
several \term{consistency models} that allow the runtime to optimize
the parallel execution while maintaining consistent computation.


\tightsubsection{Sequential Consistency Models}
\label{sec:consistency}

A parallel implementation of GraphLab must guarantee sequential
consistency \cite{Lamport:ky} over update tasks and sync operations.
We define sequential consistency in the context of the GraphLab
abstraction as:

\begin{dfn}[GraphLab Sequential Consistency]
  \label{dfn:sequential_consistency}
  For every parallel execution of the GraphLab abstraction, there
  exists a sequential ordering on all executed update tasks and sync
  operations which produces the same data graph and synced global
  values.
\end{dfn}







\begin{figure*}[t]
  \label{fig:consistency}
  \begin{center}
    \subfigure[Consistency Models]{
      \label{fig:scoperings}
      \includegraphics[width=.4\textwidth]{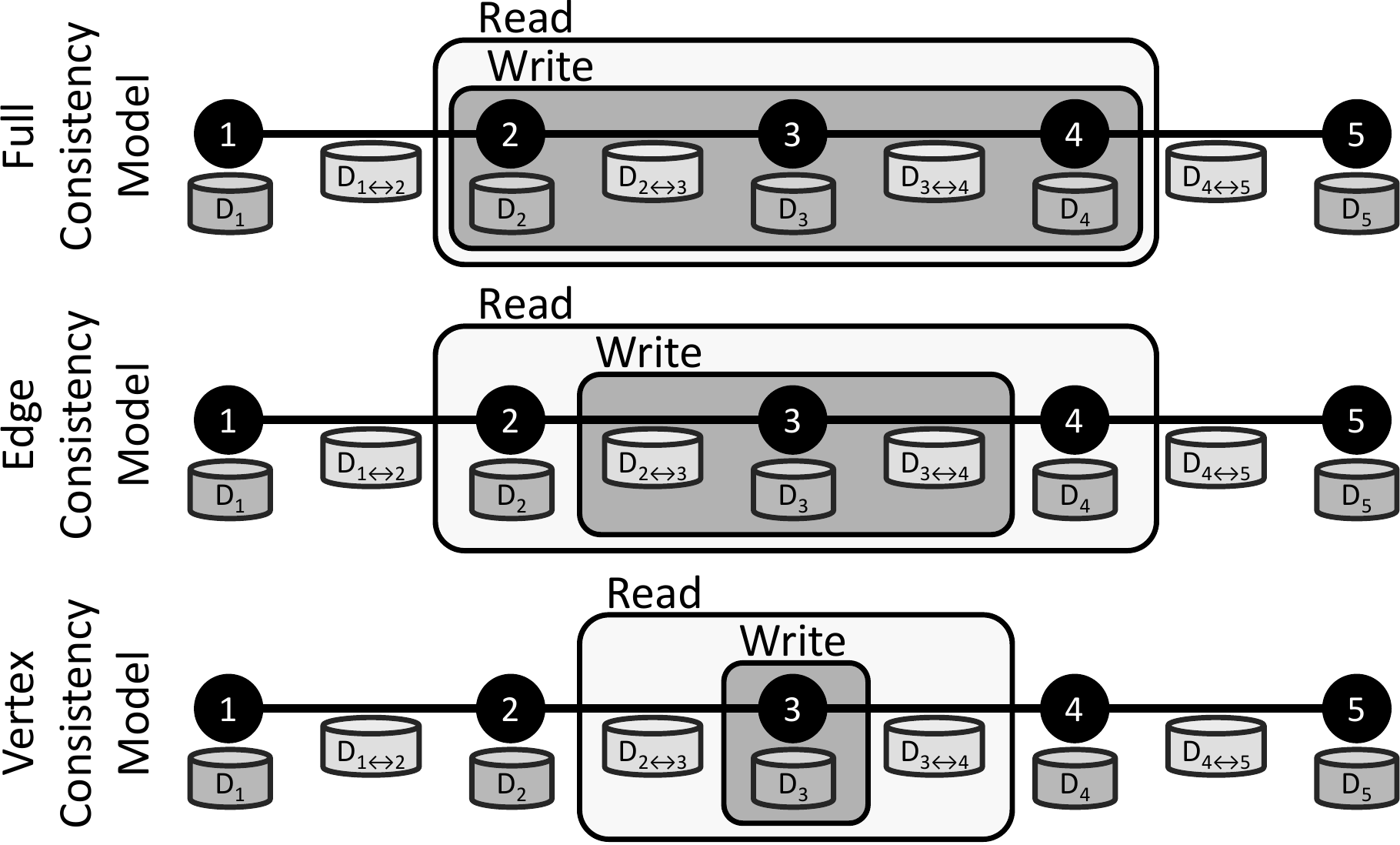}
    } \hspace{0.5cm} \subfigure[Consistency and Parallelism]{
      \label{fig:scopetradeoff}
      \includegraphics[width=.48\textwidth]{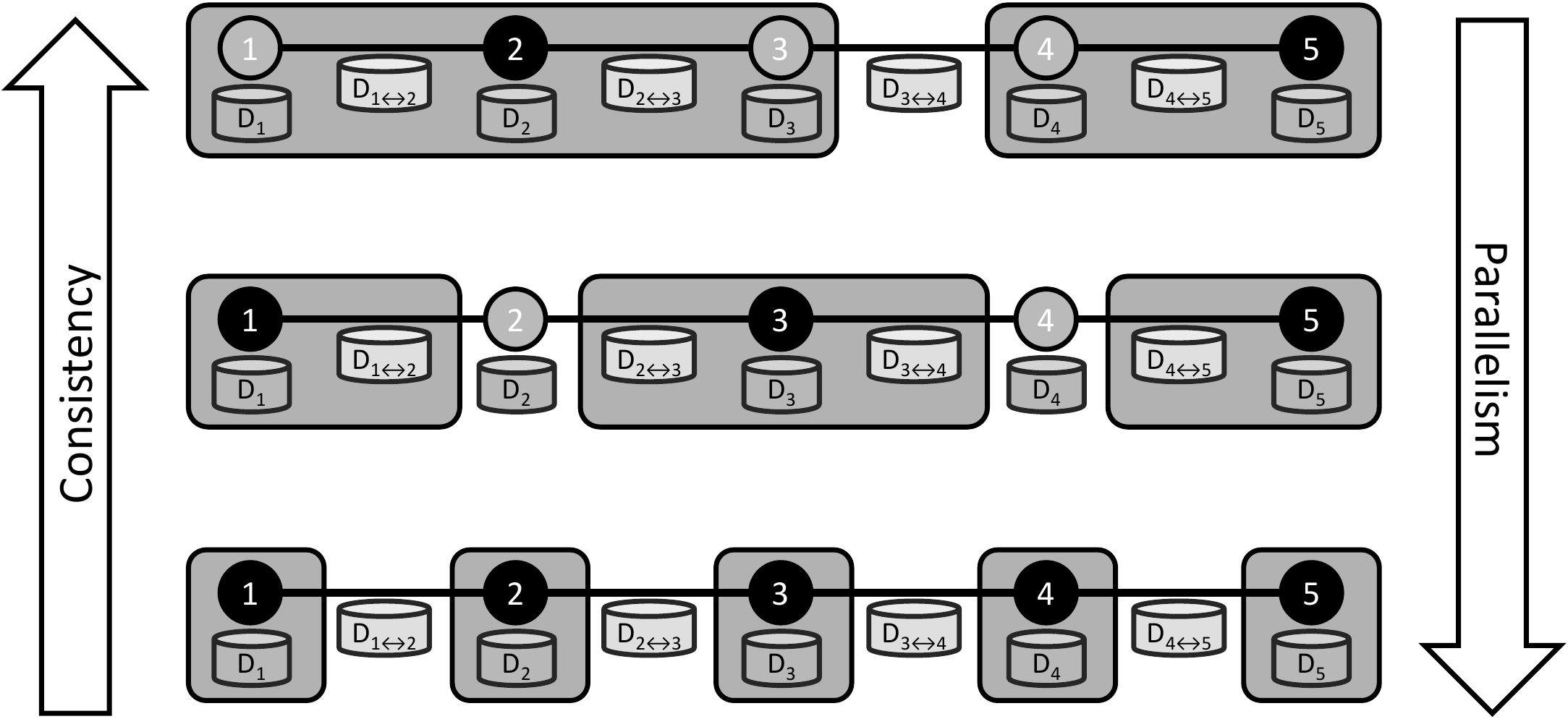}
    }
  \vspace{-2mm}
    \caption{ \footnotesize To ensure sequential consistency while
      providing the maximum parallelism, the GraphLab abstraction
      provides three different consistency models: full, edge, vertex.
      In figure \textbf{(a)}, we illustrate the read and write
      permissions for an update function executed on the central vertex
      under each of the consistency models.  
      Under the \term{full consistency model} the update function has
      complete read write access to its entire scope.  Under the
      slightly weaker \term{edge consistency model} the update
      function has only read access to adjacent vertices.  Finally,
      \term{vertex consistency model} only provides write access to
      the local vertex data.  The vertex consistency model is ideal
      for independent computation like feature processing.  In figure
      \textbf{(b)} 
      We illustrate the trade-off between consistency and
      parallelism.  The dark rectangles denote the write-locked
      regions which cannot overlap. Update functions are executed on
      the dark vertices in parallel.  Under the full consistency model
      we are only able to run two update functions $f(2, \scope{2})$
      and $f(5, \scope{5})$ simultaneously while ensuring sequential
      consistency.  Under the edge consistency model we are able to
      run three update functions (i.e., $f(1, \scope{1})$, $f(3,
      \scope{3})$, and $f(5, \scope{5})$) in parallel.  Finally under
      the vertex consistency model we are able to run update functions
      on all vertices in parallel.}
  \vspace{-6mm}
  \end{center}
 
\end{figure*}

A simple method to achieve sequential consistency among update
functions is to ensure that the scopes of concurrently executing
update functions do not overlap.  We refer to this as the \term{full
  consistency} model (see \figref{fig:scoperings}).  Full consistency
limits the potential parallelism since concurrently executing update
functions must be at least two vertices apart (see
\figref{fig:scopetradeoff}).  Even in moderately dense data graphs,
the amount of available parallelism could be low.  Depending on the
actual computation performed within the update function, additional
relaxations can be safely made to obtain more parallelism without
sacrificing sequential consistency.



We observed that for many machine learning algorithms, the update
functions do not need full read/write access to all of the data within
the scope. For instance, the PageRank update in \eqnref{eqn:pagerank}
only requires read access to edges and neighboring vertices.  To
provide greater parallelism while retaining sequential consistency, we
introduced the \term{edge consistency} model.  If the edge consistency
model is used (see \figref{fig:scoperings}), then each update function
has exclusive read-write access to its vertex and adjacent edges but
read only access to adjacent vertices.  This increases parallelism by
allowing update functions with slightly overlapping scopes to safely
run in parallel (see \figref{fig:scopetradeoff}).


Finally, for many machine learning algorithms there is often some
initial data pre-processing which only requires read access to
adjacent edges and write access to the central vertex.  For these
algorithms, we introduced the weakest \term{vertex consistency} model
(see \figref{fig:scoperings}).  This model has the highest parallelism
but only permits fully independent (\texttt{Map}) operations on vertex
data.


While sequential consistency is essential when designing,
implementing, and debugging complex ML algorithms, an adventurous user
\cite{Chafi11} may want to relax the theoretical consistency
constraints. Thus, we allow users to choose a weaker consistency model
at their own risk.




\section{Distributed GraphLab Design}
\label{sec:impldetails}
In our prior work \cite{uaigraphlab_anon} we implemented an
optimized shared memory GraphLab runtime using PThreads.  To
fully utilize clouds composed of multi-core instances, we implemented
Distributed GraphLab on top of our shared memory runtime. As
we transitioned to the distributed setting, we had to address two main 
design challenges:

\begin{list}{$\circ$}{\setlength{\leftmargin}{1em}}
\item \textbf{Distributed Graph:} To manage the data graph across
  multiple machines we needed a method to efficiently load,
  distribute, and maintain the graph data-structure over a potentially
  varying number of machines.
\item \textbf{Distributed Consistency:} To support the various
  consistency models in the distributed setting, we needed an
  efficient mechanism to ensure safe read-write access to the
  data-graph.
\end{list}

We first implemented a data graph representation that allows for rapid repartitioning
across different loads cluster sizes.
Next, we implemented two versions of the GraphLab engine for the distributed
setting, making use of asynchronous communication implemented on top
of TCP sockets. The first engine is the simpler chromatic engine
(\secref{sec:chromaticengine}) which uses basic graph coloring to
manage consistency.  The second is a locking engine
(\secref{sec:lockingengine}) which uses distributed locks.

\tightsubsection{The Distributed Data Graph}

Efficiently implementing the data graph in the distributed setting
requires balancing computation, communication, and storage.  To ensure
balanced computation and storage, each
machine must hold only a small fraction of the data graph.  At the same
time we would like to minimize the number of edges that cross partitions, to
reduce the overall state that must be synchronized across machines.
Finally, the cloud setting introduces an additional challenge.
Because the number of machines available may vary with the research budget
and the performance demands, we must be able to quickly load the
data-graph on varying sized cloud deployments. To resolve these
challenges, we developed a graph representation based on two-phased
partitioning which can be efficiently load balanced on arbitrary
cluster sizes.

\begin{figure*}[t]
  \centering
  \subfigure[Original Graph Cut]{
    \label{fig:graphcut}
    \includegraphics[width=.3\textwidth]{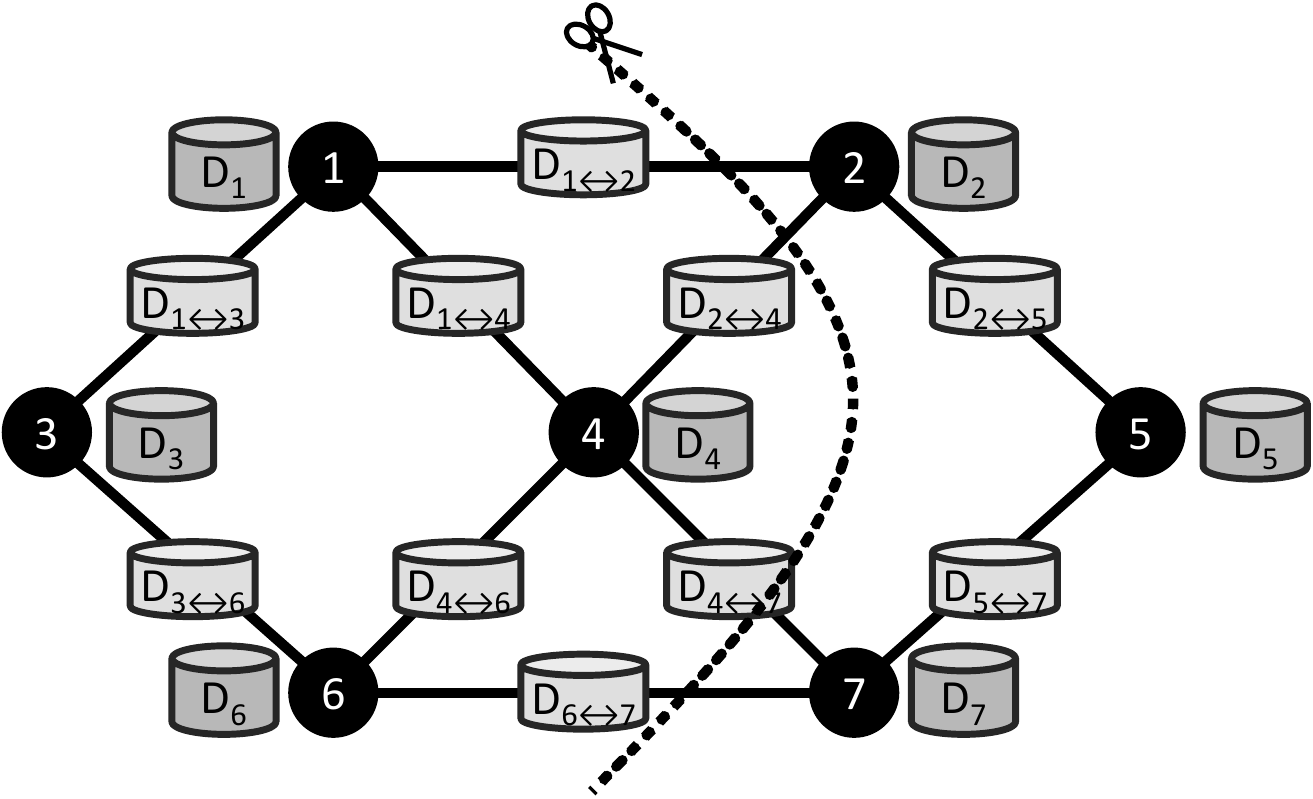} 
  }
  \subfigure[Resulting Atoms with Ghosts]{
    \label{fig:atoms}
    \includegraphics[width=.5\textwidth]{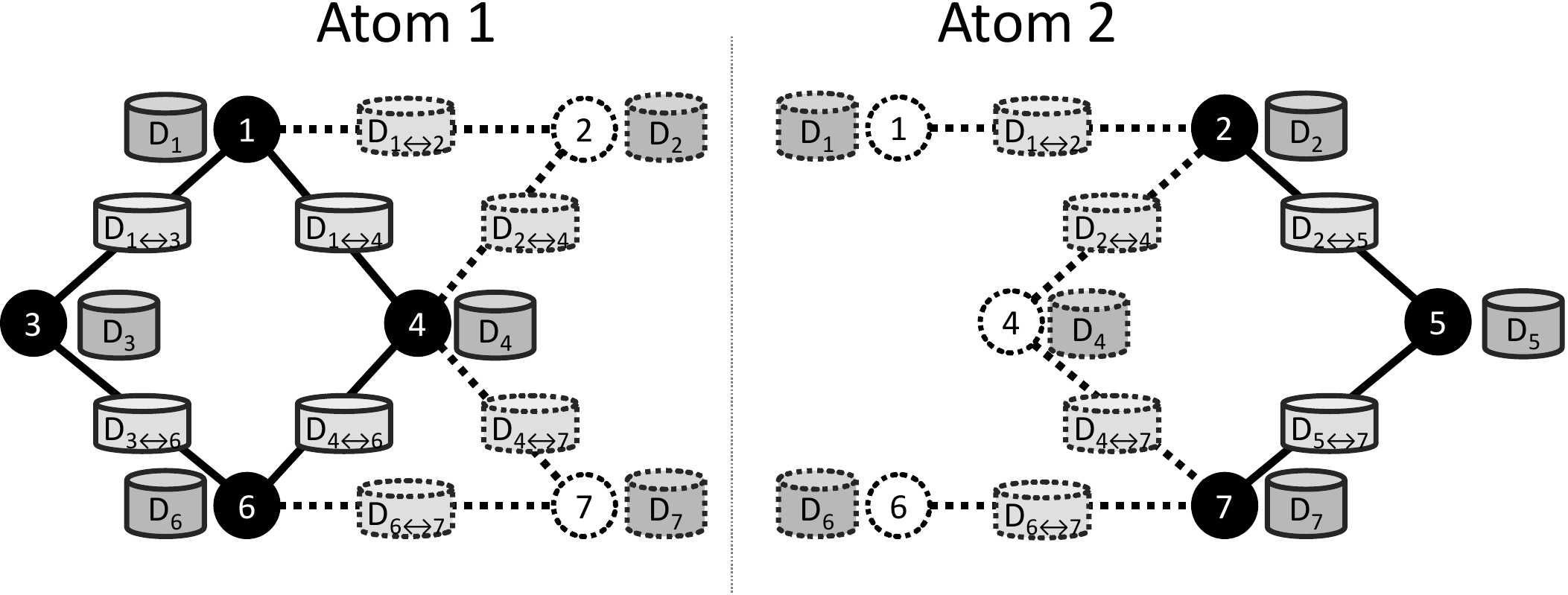}
  } 
  \subfigure[Meta-Graph]{
    \label{fig:metagraph}
    \includegraphics[width=.12\textwidth]{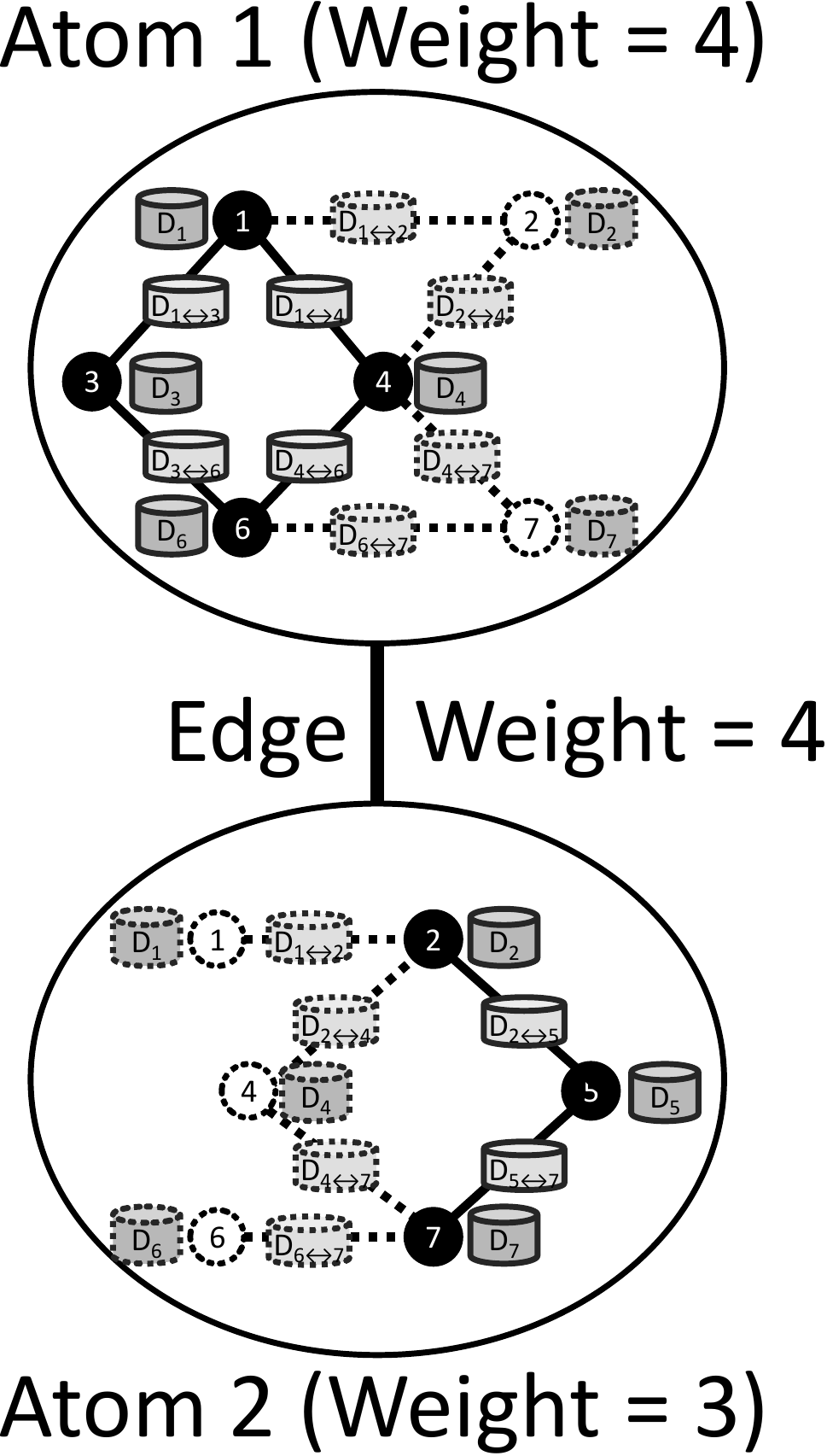}
  } 

    \vspace{-2mm}
    \caption{\footnotesize \emph{Partitions and Ghosts.} To represent
      a distributed graph we partition the graph into files each
      containing fragments of the original graph.  In \textbf{(a)} we
      cut a graph into two (unbalanced) pieces.  In \textbf{(b)} we
      illustrate the ghosting process for the resulting two atoms.
      The edges, vertices, and data illustrated with broken edges are
      \term{ghosts} in the respective atoms. The meta-graph
      \textbf{(c)} is a weighted graph of two vertices.}
    \label{fig:atomplot}
    \vspace{-6mm}
\end{figure*}

%

The graph is initially over-partitioned by an expert, or by using a
graph partitioning heuristic (for instance
Metis \cite{metis}) into $k$ parts where $k$ is much greater than
the number of machines (see \figref{fig:graphcut}). 
Each part is stored as a different file possibly on a distributed store (HDFS, Amazon S3). 
The connectivity structure of the $k$ parts is then represented as a
\term{meta-graph} with $k$ vertices. Each vertex of the \term{meta-graph} 
represents a partition, and is weighted 
by the amount of data it stores. Each edge is weighted by the 
number of edges crossing the partitions.

Distributed loading is accomplished by performing a
fast balanced partition of the meta-graph 
into $\#machines$ parts ,and each machine constructs its local partition of the
graph by merging its assigned files. To facilitate communication,
each machine also stores the \term{ghost} of its local partition:  
the set of vertices and edges adjacent to the partition boundary (see \figref{fig:atoms}).
The ghosts conveniently act as local caches for their true counterparts across the network,
and cache coherence is managed using versioning \cite{Bernstein1981}.


The two-stage partitioning technique allows one graph partition to be
reused for different numbers of machines without incurring a repartitioning step. 
A study on the quality of the two stage partitioning scheme is beyond the scope of 
this paper, though simple experiments using graphs obtained from 
\cite{stanfordlargenetwork} suggest that the performance is comparable to direct partitioning.

\tightsubsection{GraphLab Engines}

The GraphLab \term{engine} emulates the \emph{execution model} defined
in \secref{sec:execution_model} and is responsible for executing
update task and sync operations, maintaining the update task set
$\mathcal{T}$, and ensuring sequential consistency with respect to the
appropriate consistency model (see \secref{sec:consistency}).  As
discussed earlier in \secref{sec:execution_model}, variations in how
$\mathcal{T}$ is maintained and the order in which elements are
removed is up to the implementation and can affect performance and
expressiveness.  To evaluate this trade-off we built the low-overhead
\term{Chromatic Engine}, which executes $\mathcal{T}$ in a fixed
order, and the more expressive \term{Locking Engine} which executes
$\mathcal{T}$ in an asynchronous prioritized order.


\subsubsection{Chromatic Engine}
\label{sec:chromaticengine}



The \term{Chromatic Engine} imposes a static ordering on the update
task set $\mathcal{T}$ by executing update task in a canonical order
(e.g., the order of their vertices).  A classic technique
\cite{bertsekas} to achieve a sequentially consistent parallel
execution of a set of dependent tasks is to construct a vertex
coloring.  A vertex coloring assigns a color to each vertex such that
no adjacent vertices share the same color.  Given a vertex color of
the data graph, we can satisfy the \emph{edge consistency model} by
executing, \emph{in parallel}, all update tasks with vertices of the
same color before proceeding to the next color.  The sync operation
can be run safely between colors.

We can satisfy the other consistency models simply by changing how the
vertices are colored.  The stronger \emph{full consistency model} is
satisfied by constructing a second-order vertex coloring (i.e., no
vertex shares the same color as any of its distance two neighbors).
Conversely, we can trivially satisfy the \emph{vertex consistency
  model} by assigning all vertices the same color.

In the distributed setting it is necessary to both prevent
overlapping computation, and also synchronize any changes to ghost
vertices or edge data between colors.  A full communication barrier is
enforced between color phases to ensure completion of all data
synchronization before the next color begins. To maximize network
throughput and to minimize time spent in the barrier, synchronization
of modified data is constantly performed in the background while
update functions are executing.



Approximate graph coloring can be quickly obtained through graph
coloring heuristics.  Furthermore, many ML problems have obvious
colorings. For example,
many optimization problems in ML are naturally expressed as bipartite
graphs (\secref{sec:applications}), while problems based upon
templated Bayesian Networks \cite{friedmankoller} can be easily
colored by the expert through inspection of the template model
\cite{gonzalez11}.

The simple design of the Chromatic engine is made possible by the 
explicit communication structure defined by the data graph, allowing
data to be pushed directly to the machines requiring the information. In addition,
the cache versioning mechanism further optimizes communication
by only transmitting modified data.
The advantage of the Chromatic engine lies its predictable execution
schedule. Repeated invocations of the chromatic engine will always
produce identical update sequences, regardless of the number of
machines used.  This property makes the Chromatic engine highly
suitable for testing and debugging purposes.
We provide a distributed debugging tool which halts at the end of each
color, allowing graph data and scheduler state to be queried.


\subsubsection{Locking Engine}
\label{sec:lockingengine}
Even though the chromatic engine is a complete implementation of the
GraphLab abstraction as defined in \secref{sec:graphlababstraction},
it does not provide sufficient scheduling flexibility for many
interesting applications. Here we describe an implementation which
directly extends from a typical shared memory implementation to the
distributed case.

In the shared memory implementation of GraphLab, the consistency
models were implemented by associating a readers-writer lock with each
vertex.  The vertex consistency model is achieved by acquiring a write
lock on the central vertex of each requested scope. The
edge-consistency model is achieved by acquiring a write lock on the
central vertex, and read locks on adjacent vertices. Finally, full
consistency is achieved by acquiring write locks on the central vertex
and all adjacent vertices.

The main execution loop in the shared memory setting uses worker
threads to pull tasks from the scheduler, acquire the required locks,
evaluate the task, and then release the locks.  This loop is repeated
until the scheduler is empty.  The sync operation is triggered by a
global shared-memory task counter.  Periodically, as sync operations
become ready, all threads are forced to synchronize in a barrier to
execute the sync operation.

In the distributed setting, the same procedure is used. However, since
the graph is partitioned, we restrict each machine to only run updates
on vertices it owns.  The ghost vertices/edges ensure that the update
will always have direct memory access to all information in the scope,
and distributed locks are used to ensure that the ghost is up to date.
Finaly, the scheduling flexibility permitted by the abstraction allow the
use of efficient approximate FIFO/priority task-queues.
Distributed termination is evaluated using a multi-threaded variant of
the distributed consensus algorithm described in \cite{Misra83}.


Since the distributed locking and synchronization introduces
substantial latency, we rely on several techniques to reduce latency
and hide its effects \cite{gupta91}.  First, the ghosting system
provides caching capabilities eliminating the need to wait on data
that has not changed remotely.  Second, all locking requests 
and synchronization calls are \emph{pipelined} allowing each thread to
request multiple scope locks simultaneously and then evaluate the
update tasks only when the locks are satisfied. The \emph{lock pipelining}
technique we implemented shares similarities to the continuation 
passing method in \cite{Turek92}.

\tightsection{Applications}
\label{sec:applications}


To evaluate the performance of the distributed GraphLab runtime as
well as the representational capabilities of the GraphLab abstraction,
we implemented several state-of-the-art ML algorithms.  Each algorithm
is derived from active research in machine learning and is applied to
real world data sets with different structures (see
\tableref{table:appinp}), update functions, and sync operations.  In
addition, each application tests different features 
of the distributed
GraphLab framework.
The source and datasets for all the applications may be obtained from
[\textbf{http://graphlab.org}].

\begin{table*}[t]
\begin{small}
  \begin{center}
  \begin{tabular}{|c|c|c|C{3.5em}|C{3em}|C{6em}|c|c|c|}
    \hline
    \textbf{Exp.} &
    \textbf{\#Verts} & 
    \textbf{\#Edges} &
    \textbf{Vertex Data} &
    \textbf{Edge Data} &
    \textbf{Update Complexity} &
    \textbf{Shape} & 
    \textbf{Partition} & 
    \textbf{Engine} \\
    \hline
    \hline
    Netflix & 
    0.5M & 
    99M & 
    $8 d + 13$  &
    $16$  &
    $\BigO{d^3 + deg.}$ &
    bipartite & 
    random & 
    Chromatic\\
    \hline
    CoSeg &
    10.5M & 
    31M & 
    $392$  &
    $80$  &
    $\BigO{deg.}$ &
    3D grid & 
    frames & 
    Locking\\
    \hline
    NER & 
    2M & 
    200M & 
    $816$  &
    $4$  &
    $\BigO{deg.}$ &
    bipartite & 
    random & 
    Chromatic\\
    \hline
  \end{tabular}
  \vspace{-2mm}
  \caption{\footnotesize {\emph{Experiment input sizes.} The vertex and edge data are
    measured in bytes.} }
  \vspace{-3mm}
  \label{table:appinp}
  \end{center}
 \end{small}
\end{table*}

\begin{figure*}[t]
  \vspace{-2mm}
  \centering
\subfigure[Netflix Prediction Error]{
  \label{fig:alsaccuracy}
    \includegraphics[width=.23\textwidth]{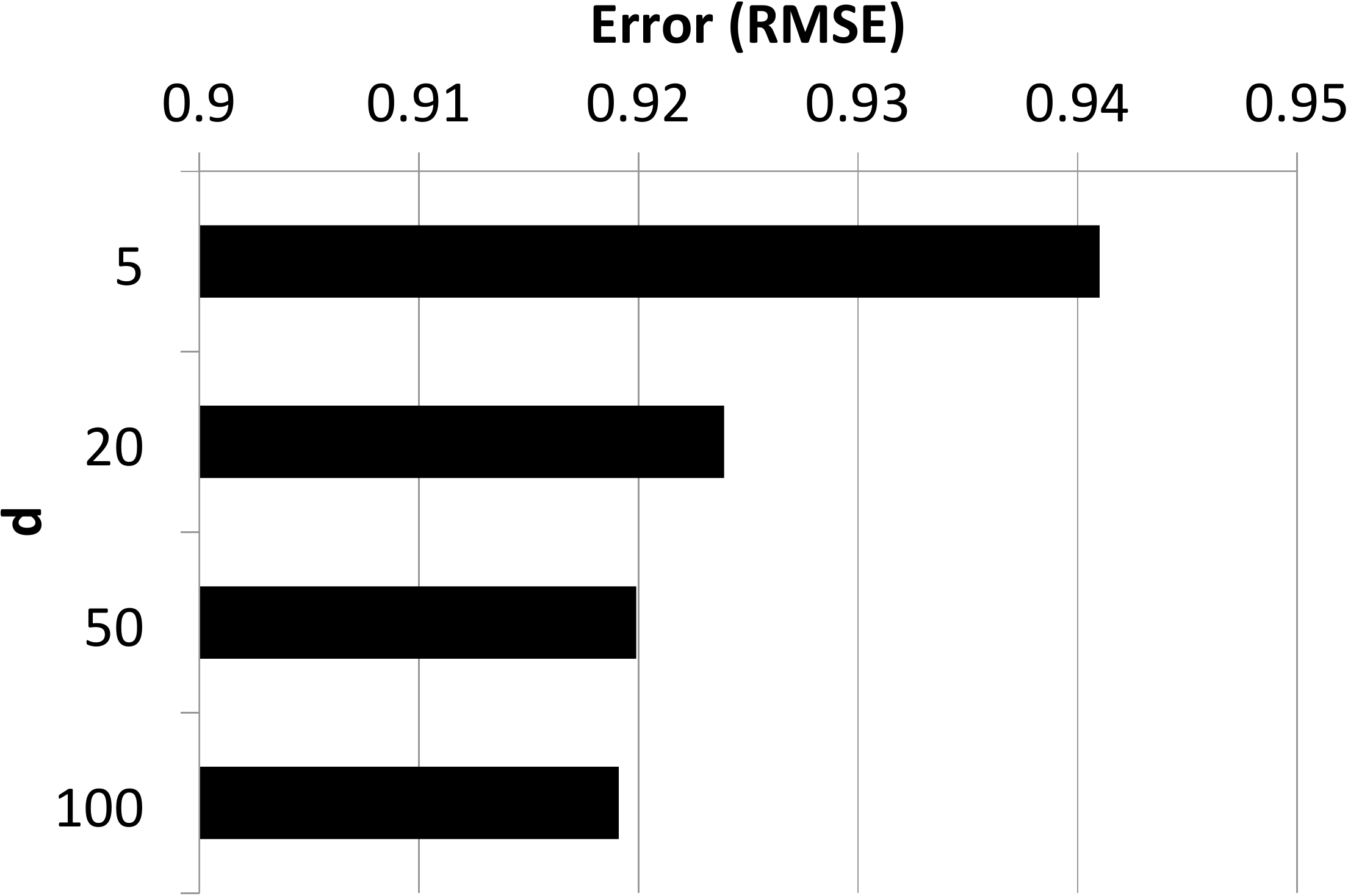}
}
  \subfigure[Video Frame]{
    \label{fig:videoframe}
    \includegraphics[width=.18\textwidth]{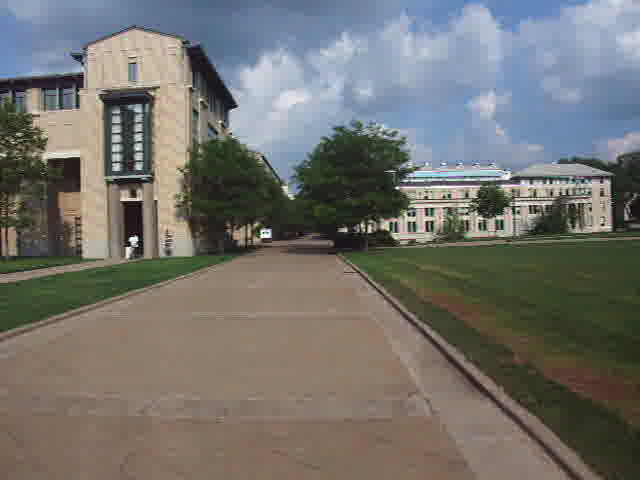}
  }
  \subfigure[Segmentation]{
    \label{fig:cosegframe}
    \includegraphics[width=.18\textwidth]{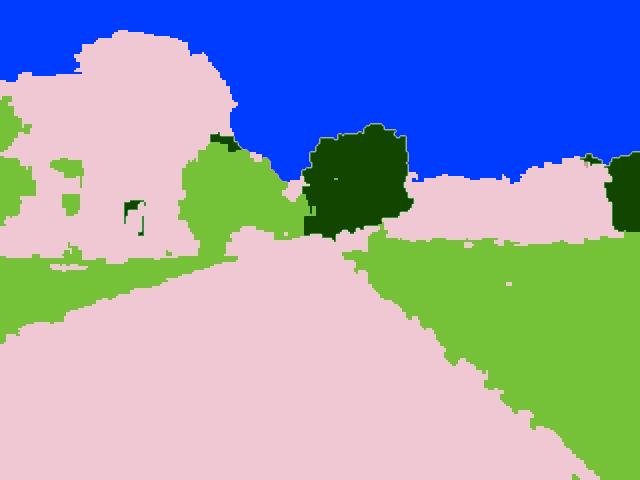}
  }
  \subfigure[NER] {
                \footnotesize
                \raisebox{10mm}{
      \begin{tabular}{|c|c|c|c|}
      \hline
      \textbf{Food} &  \textbf{Religion} & \textbf{City} \\
      \hline \hline
      onion & Catholic    & Munich \\
      garlic & Fremasonry  & Cape Twn.   \\
      noodles & Marxism  & Seoul \\
      blueberries & Catholic Chr. & Mexico Cty. \\
      beans & Humanism & Winnipeg \\
      \hline 
    \end{tabular}     
    \label{tbl:coemwords}
    }
  }
  \vspace{-3mm}
  \caption{ \footnotesize \textbf{(a) Netflix: } The test error (RMSE)
    of the ALS algorithm on the Netflix dataset after 30 iterations
    with different values of $d$. Lower RMSE is more
    accurate. 
    \textbf{(b) Coseg:} a sample from the original video
    sequences. \textbf{(c) Coseg:} results of running the
    co-segmentation algorithm. We observe that the algorithm
    successfully identified the common segments such as ``sky'' and
    ``grass.'' \textbf{(d) NER: } Top words for several types. }
  \vspace{-3mm}
\end{figure*}

\tightsubsection{Netflix Movie Recommendation} 
\label{sec:netflix}
The Netflix movie recommendation task \cite{Netflix} uses
\emph{collaborative filtering} to predict the movie ratings of users,
based on the ratings of similar users.  The \term{alternating least
  squares} (\term{ALS}) algorithm is commonly used in collaborative
filtering. The input to ALS is a sparse users by movies matrix $R,$
containing the movie ratings of each user. The algorithm proceeds by
computing a low rank approximate matrix factorization:
\begin{equation*}
  \label{eqn:als}
  \includegraphics[width=.4\textwidth]{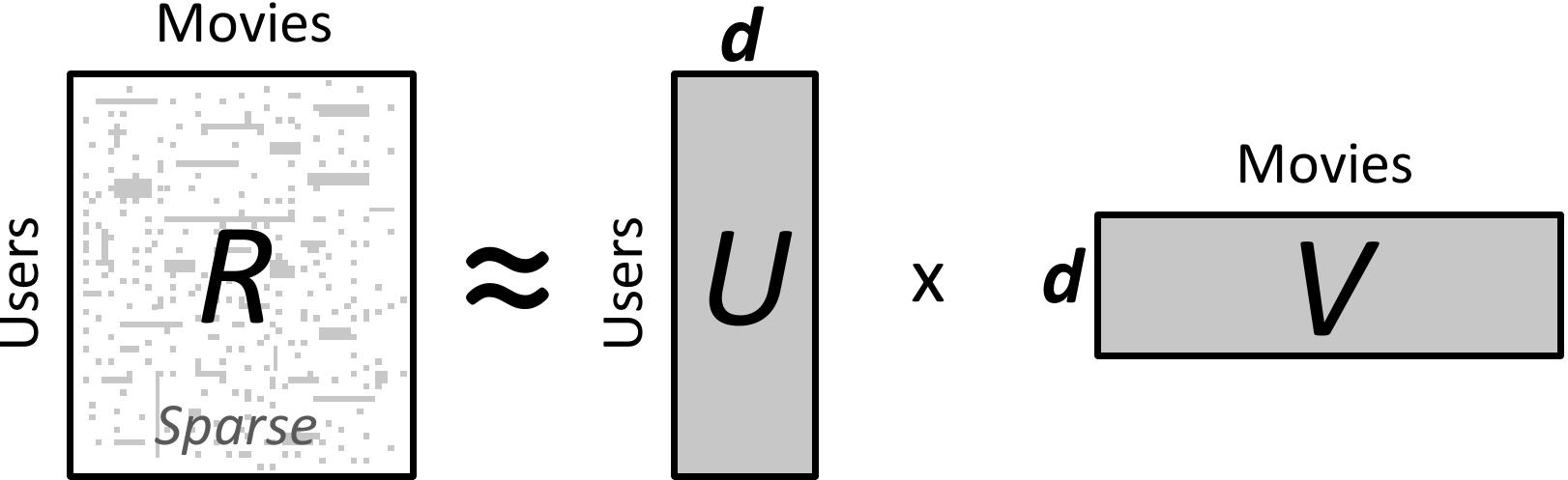}
\end{equation*}
where $U$ and $V$ are rank $d$ matrices.  The ALS algorithm alternates
between computing the least-squares approximation for $U$ or $V$ while
holding the other fixed.  The quality of the approximation depends on the magnitude of $d$, as shown in
\figref{fig:alsaccuracy}. 



While ALS may not seem like a graph algorithm, it can be represented
elegantly using the GraphLab abstraction.    The \emph{sparse} matrix $R$
defines a bipartite graph (see \tableref{table:appinp}) connecting
each user with the movie he/she rated.  The edge data contains the rating
for a movie-user pair.  The vertex data for users and movies contains 
the corresponding row in $U$ and column in $V$ respectively.  

The ALS algorithm can be encoded as an update function that recomputes
the least-squares solution for the current movie or user given the
neighboring users or movies.  Each local computation is accomplished
using BLAS/LAPACK linear algebra library for efficient matrix
operations.  
The bipartite graph is naturally two colored, thus the program is
executed using the chromatic engine with two colors.  A sync operation
is used to compute the prediction error during the run. Due to the
density of the graph, a random partitioning was used.

 
\tightsubsection{Video Cosegmentation (CoSeg)}
\label{sec:coseg}
Video cosegmentation automatically identifies and clusters
spatio-temporal segments of video (see \figref{fig:videoframe}) that
share similar texture and color characteristics.  The resulting
segmentation (see \figref{fig:cosegframe}) can be used in scene
understanding and other computer vision and robotics applications.
Previous cosegmentation methods \cite{batra-icoseg} have 
focused on processing frames in isolation.  As part of this work,
we developed a joint cosegmentation algorithm that processes 
all frames simultaneously and therefore is able to model 
\emph{temporal} stability.

We preprocessed $1,740$ frames of high-resolution video by coarsening
each frame to a regular grid of $120 \times 50$ rectangular
\term{super-pixels}. Each super-pixel stores the color and texture
statistics for all the raw pixels in its domain.  The CoSeg algorithm
predicts the best label (e.g., sky, building, grass, pavement, trees)
for each super pixel using \term{Gaussian Mixture Model} (\term{GMM})
\cite{bishop} in conjunction with \term{Loopy Belief Propagation}
(\term{LBP}) \cite{pearl88}. The GMM estimates the best label given
the color and texture statistics in the super-pixel.  The algorithm
operates by connecting neighboring pixels in time and space into a
large three-dimensional grid and uses LBP to smooth the local
estimates.  We combined the two algorithms so that CoSeg alternates
between LBP to compute the label for each super-pixel given the
current GMM and then updating the GMM given the labels from LBP.

The GraphLab data graph structure for video cosegmentation is the
three dimensional grid graph (see \tableref{table:appinp}). 
The vertex data stores the current label distribution as well as the
color and texture statistics for each super-pixel.  The edge data
stores the parameters needed for the LBP algorithm. The parameters for
the GMM are maintained using the sync operation.
The GraphLab update function executes the LBP local iterative update. 
 We implement the state-of-the-art adaptive update schedule
described by \cite{elidan06}, where updates which are expected to change
vertex values significantly are prioritized. 
We therefore make use of the locking engine with the 
approximate priority ordering task queue.  Furthermore, the
graph has a natural partitioning by slicing across frames. This also
allows feature processing of the video to be performed in an 
embarrassingly parallel fashion, permitting the use of Hadoop for 
preprocessing.


\tightsubsection{Named Entity Recognition (NER)}
\label{sec:coem}
Named Entity Recognition (NER) is the task of determining the type
(e.g., \texttt{Person}, \texttt{Place}, or \texttt{Thing}) of a
\term{noun-phrase} (e.g., \textit{Obama}, \textit{Chicago}, or
\textit{Car}) from its \term{context} (e.g., \textit{``President \_\_
  \ldots''}, \textit{``\ldots lives near \_\_.''}, or \textit{``\ldots
  bought a \_\_.''}).  NER is used in many natural language processing
applications as well as information retrieval.  In this application we
obtained  a large crawl of the web 
and we counted the number of occurrences of each noun-phrase in each context.  
Starting with a small seed set of pre-labeled noun-phrases,  the CoEM algorithm  \cite{CoEMJones}
labels the remaining noun-phrases and contexts (see
\tableref{tbl:coemwords}) by alternating
between estimating the best assignment to each noun-phrase given the
types of its contexts and estimating the type of each context given
the types of its noun-phrases.

The GraphLab data graph for the NER
problem is bipartite with vertices corresponding to each
noun-phrase on one side and vertices corresponding to each context
on the other.  There is an edge between a noun-phrase and a context if
the noun-phrase occurs in the context.  The vertex for both
noun-phrases and contexts stores the estimated distribution over
types.  The edge stores the number of times the noun-phrase
appears in that context.

The NER computation is represented in a simple GraphLab update
function which computes a weighted sum of probability tables stored on adjacent
vertices and then normalizes.  
Once again, the bipartite graph is naturally two colored, allowing
us to use the chromatic scheduler. Due to the
density of the graph, a random partitioning was used. Since the NER 
computation is relatively light weight and uses only simple floating 
point arithmetic; combined with the use of a random partitioning, this
application stresses the overhead of the GraphLab runtime 
as well as the network.
 
\tightsubsection{Other Applications}
\label{sec:otherapplications}
In the course of our research, we have also implemented several
other algorithms, which we describe briefly:

\textbf{Gibbs Sampling on a Markov Random Field.} The task is to
  compute a probability distribution for a graph of random variables
  by sampling. Algorithm proceeds by sampling a new value for each
  variable in turn conditioned on the assignments of the neighboring
  variables. Strict sequential consistency is necessary to preserve statistical properties \cite{gonzalez11}.
  
\textbf{Bayesian Probabilistic Tensor Factorization (BPTF).} This
  is a probabilistic Markov-Chain Monte Carlo version of Alternative Least Squares that also
  incorporates time-factor into the prediction. In this case, the
  tensor $R$ is decomposed into three matrices $R \approx\ V
  \bigotimes U \bigotimes T$ which can be represented in GraphLab as a 
  tripartite graph. 

In addition, GraphLab has been used successfully in several other 
research projects like clustering communities in the twitter network, collaborative filtering for
BBC TV data as well as non-parametric Bayesian inference.


\tightsection{Evaluation}
\label{sec:experiments}
We evaluated GraphLab on the three applications (Netflix, CoSeg and
NER) described above using important large-scale real-world problems
(see \tableref{table:appinp}).  We used the Chromatic engine for the
Netflix and NER problems and the Locking Engine for the CoSeg
application. Equivalent Hadoop and MPI implementations were also
tested for both the Netflix and the NER application.  An MPI
implementation of the asynchronous prioritized LBP algorithm needed
for CoSeg requires building an entirely new asynchronous sequentially
consistent system and is beyond the scope of this work.

Experiments were performed on Amazon's Elastic Computing Cloud (EC2)
using up to 64 High-Performance Cluster (HPC) instances
(\texttt{cc1.4xlarge}).  The HPC instances (as of February 2011) have
2 x Intel Xeon X5570 quad-core Nehalem architecture with 22 GB of
memory, connected by a low latency 10 GigaBit Ethernet network. All
our timings include data loading time and are averaged over three or
more runs.  Our principal findings are:
\begin{list}{$\circ$}{\setlength{\leftmargin}{1em}}
\item \emph{GraphLab is fast!} On equivalent tasks, GraphLab
  outperforms Hadoop by 20x-60x and is as fast as custom-tailored MPI
  implementations.
\item GraphLab's performance scaling improves with higher computation
  to communication ratios. When communication requirements are high,
  GraphLab can saturate the network, limiting scalability.
\item The GraphLab abstraction more compactly expresses the Netflix,
  NER and Coseg algorithms than MapReduce or MPI.
\end{list} 

\begin{figure*}[t]
\centering
\subfigure[Overall Scalability]{
    \includegraphics[width=.23\textwidth]{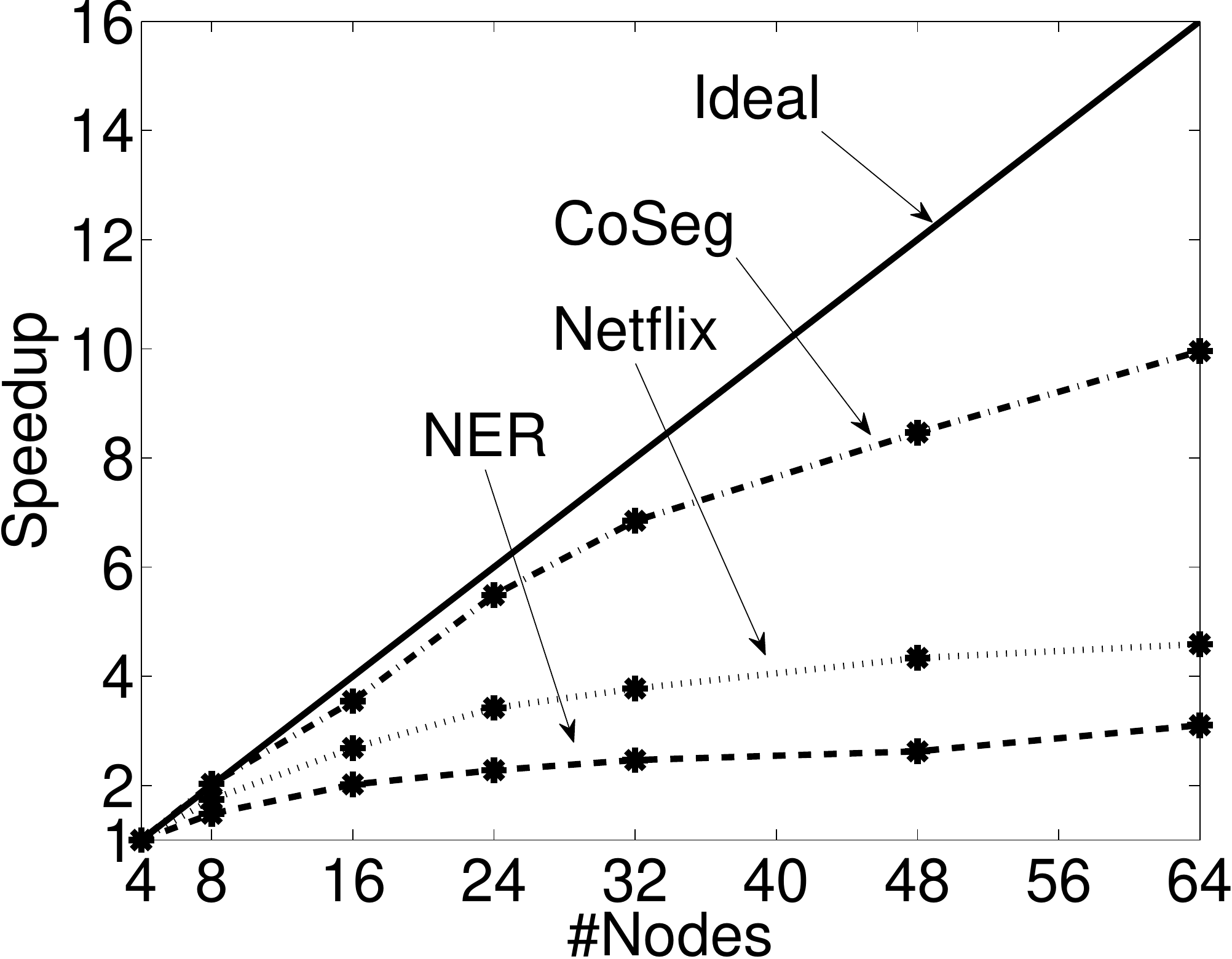}
  \label{fig:speedupmain}
}
\subfigure[Overall Network Utilization]{
    \includegraphics[width=.23\textwidth]{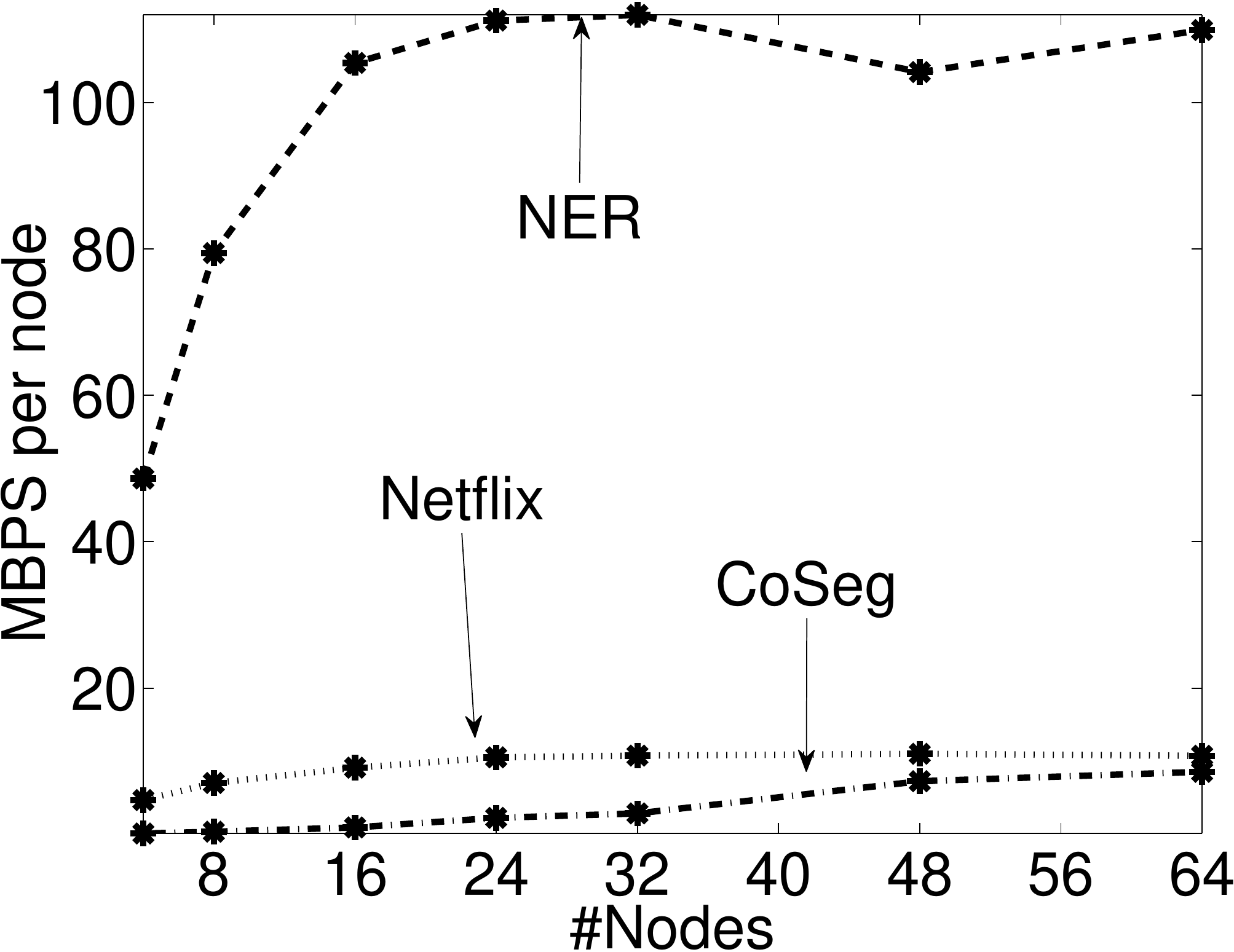}
  \label{fig:networkpernode}
}
\subfigure[Netflix Scaling with Intensity] {
    \includegraphics[width=.23\textwidth]{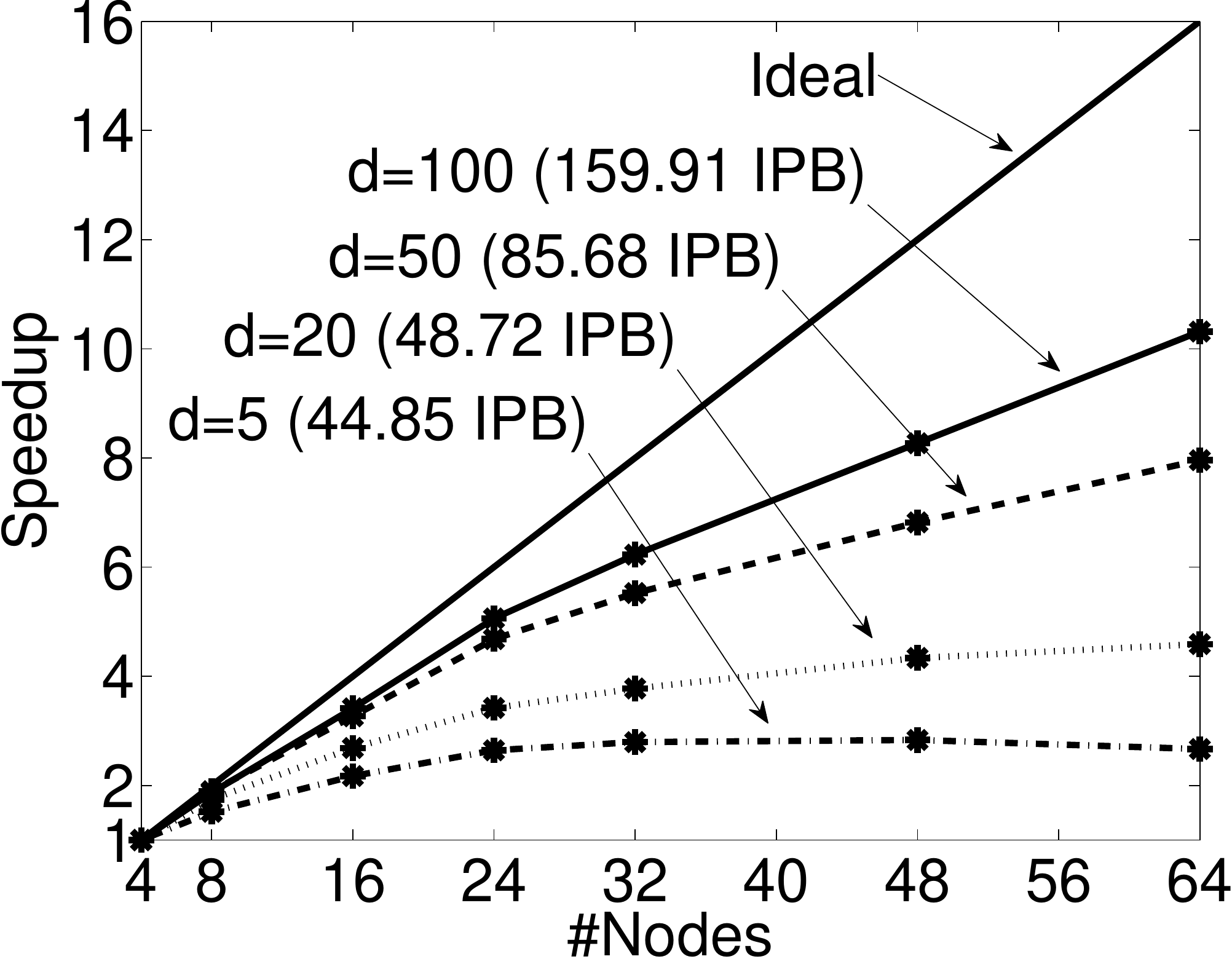}
  \label{fig:compcomplex}
}
\subfigure[Netflix Comparisons] {
    \includegraphics[width=.23\textwidth]{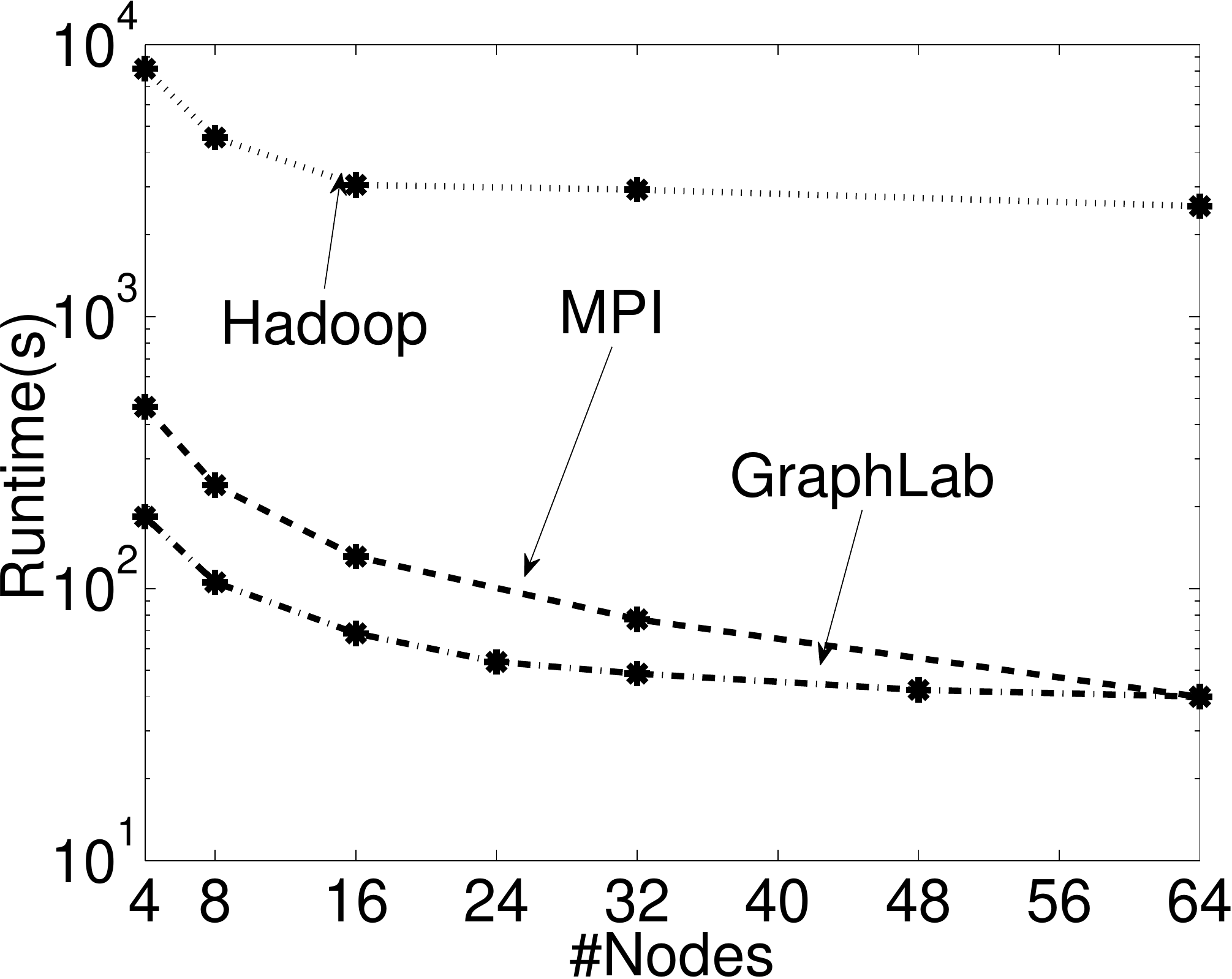}
  \label{fig:alshadoop}
}
\vspace{-4mm}
\caption{\footnotesize \textbf{(a)} Scalability of the three example
  applications with the default input size.  CoSeg scales excellently
  due to very sparse graph and high computational intensity. Netflix
  with default input size scales moderately while NER is hindered by
  high network utilization. See \secref{sec:experiments} for a
  detailed discussion. \textbf{(b)} Average number of megabytes sent
  per cluster node per second. Netflix and CoSeg have very low
  bandwidth requirements while NER appears to saturate the network
  when \#nodes increases above 24.  \textbf{(c)} Scalability of
  Netflix when computational intensity is varied. IPB refers to the
  average number of instructions per byte of data accessed by the
  update function. Increasing computational intensity improves
  parallel scalability quickly.  \textbf{(d)} Runtime of the Netflix
  experiment with GraphLab, Hadoop and MPI implementations. Note the
  logarithmic scale. GraphLab outperforms Hadoop by over 20-30x and is
  comparable to an MPI implementation. See \secref{sec:netflixandner}
  for a detailed discussion.}
\vspace{-4mm}
\end{figure*}

\tightsubsection{Scaling Performance}
\label{sec:scaling}
In \figref{fig:speedupmain} we present the parallel speedup of
GraphLab when run on 4 to 64 HPC nodes.  Speedup is measured relative
to the 4 HPC node running time.  On each node, GraphLab spawned eight
shared memory engine threads (matching the number of cores).  Of the
three applications, CoSeg demonstrated the best parallel speedup,
achieving nearly ideal scaling up to 32 nodes and moderate scaling up
to 64 nodes.  The excellent scaling of the CoSeg application can be
attributed to its sparse data graph (maximum degree 6) and a
computationally intensive update function.  While Netflix demonstrated
reasonable scaling up to 16 nodes, NER achieved only modest 3x
improvement beyond 16x or more nodes.


We attribute the poor scaling performance of NER to the large vertex
data size (816 bytes), dense connectivity, and poor partitioning
(random cut) which resulted in substantial communication overhead per
iteration.  \figref{fig:networkpernode} shows for each application,
the average number of bytes transmitted by each node per second as the
cluster size is increased.
Beyond 16 nodes, it is evident that NER saturates the network, with
each machine sending at a rate of over 100MB per second. Note that
\figref{fig:networkpernode} plots the \emph{average} transmission rate
and the peak rate could be significantly higher.



To better understand the effect of the ratio between computation and
communication on the scaling of the GraphLab abstraction, we varied the
computational complexity of the Netflix experiment.  The amount of
computation performed in Netflix can be controlled by varying $d$: the
dimensionality of the approximating matrices in \eqnref{eqn:als}.  In
\figref{fig:compcomplex} we plot the speedup achieved for varying
values of $d$ and the corresponding number of \emph{instructions per
  byte} (IPB) of data accessed.  The speedup at 64 nodes increases
quickly with increasing IPB indicating that speedup is strongly
coupled. 




\tightsubsection{Comparison to Other Frameworks}
\label{sec:netflixandner}

In this section, we compare our GraphLab implementation of the Netflix 
application and the NER application to an algorithmically equivalent Hadoop 
and MPI implementations.

\textbf{MapReduce/Hadoop.} We chose to compare against Hadoop, due to
its wide acceptance in the Machine Learning community for large scale
computation (for example the Mahout project \cite{cheng06,pegasus}). 
Fair comparison is difficult since
Hadoop is implemented in Java while ours is
highly optimized C++. Additionally, to enable fault tolerance, 
Hadoop   stores interim results to disk.  In our experiments, 
to maximize Hadoop's performance, we reduced the 
Hadoop Distributed Filesystem's (HDFS) replication factor to one,
eliminating fault tolerance. A significant amount of our effort was  spent
tuning the Hadoop job parameters to improve performance. 

\figref{fig:alshadoop} shows the running time for one iteration of
Netflix application on GraphLab, Hadoop and MPI ($d=20$ for all
cases), using between 4 and 64 nodes.  The Hadoop evaluation makes use
of the of Sebastian Schelter contribution to the Mahout
project\footnote{https://issues.apache.org/jira/browse/MAHOUT-542},
while the MPI implementation was written from scratch.  We find that
GraphLab performs the same computation between \textbf{40x-60x} times
faster than Hadoop.

\begin{figure}[t]
        \centering
        \subfigure[NER Comparisons] {
                \includegraphics[width=.22\textwidth]{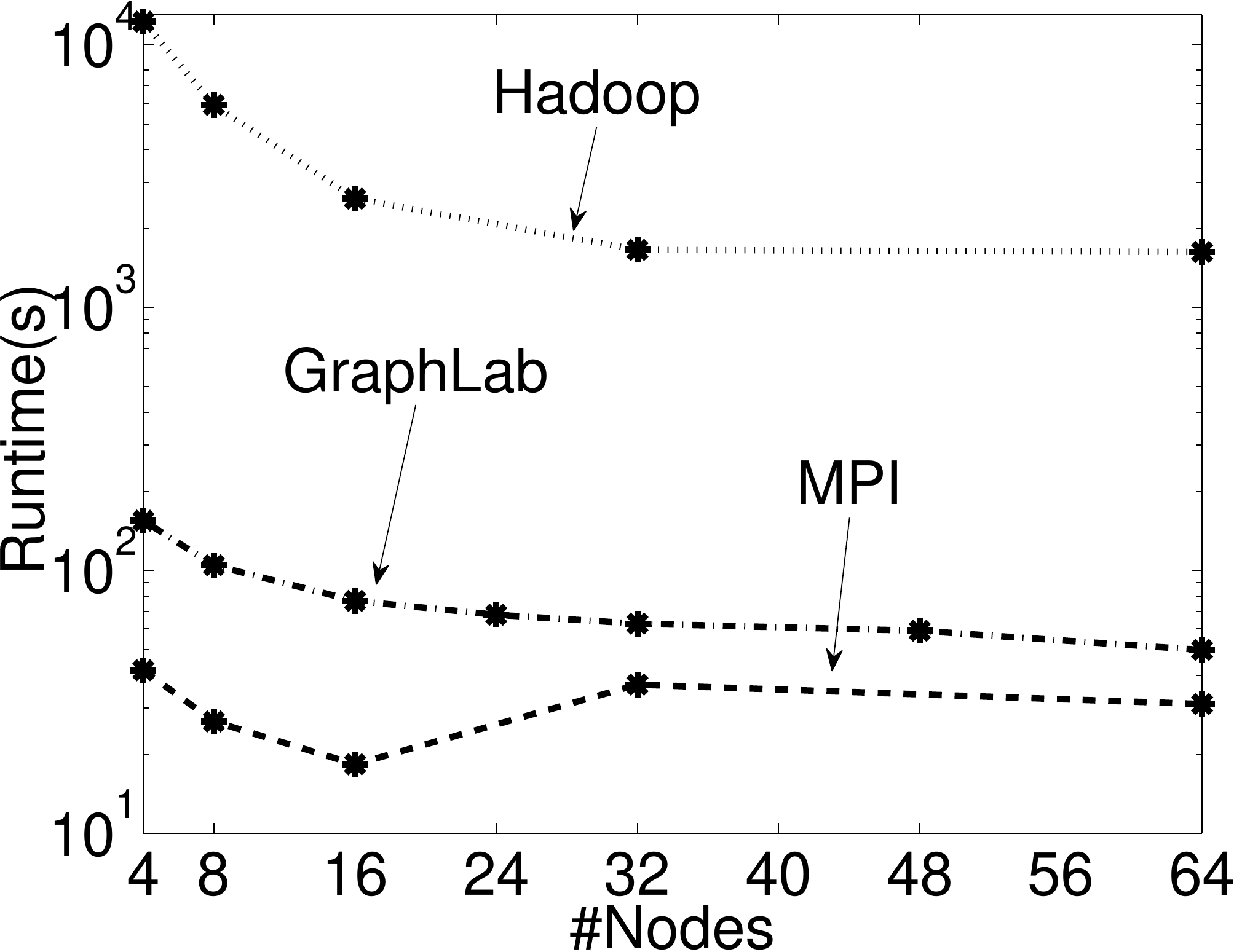}
                \label{fig:coemhadoop}
        }
    \vspace{-3mm}
    \caption{\footnotesize \textbf{(a)} Runtime of the NER experiment
     with GraphLab, Hadoop and MPI implementations. Note the logarithmic
      scale. GraphLab outperforms Hadoop by about 80x when the number of nodes is small,
      and about 30x when the number of nodes is large. The performance of 
      GraphLab is comparable to the MPI implementation. 
      }
\end{figure}

\figref{fig:coemhadoop} plots the running for one iteration of the NER application
on GraphLab, Hadoop and MPI. The Hadoop implementation was aggressively optimized:
we implemented specialized binary marshaling methods which improve
performance by  5x over a baseline implementation.
\figref{fig:coemhadoop} shows that the 
GraphLab implementation of NER  obtains a 20-30x speedup over Hadoop.  

Part of the performance of GraphLab over Hadoop
 can be explained by implementation differences, 
but it is also easy to see that the GraphLab representation of
both NER and ALS is inherently more efficient. For instance the case of 
NER, when implemented in Hadoop, the
Map-function, normally the cornerstone of embarrassing-parallelism in MapReduce 
essentially does no work. The Map only serves to emit the vertex probability table
for every edge in the graph, which corresponds to over 100 gigabytes of HDFS 
writes occurring between the Map and Reduce stage. The cost of this operation
can easily be multiplied by factor of three if HDFS replication is turned on. 
Comparatively, the GraphLab update function is simpler 
as users do not need to explicitly define the flow of information from the Map to the Reduce,
but simply modifies data in-place. In the case of such iterative computation,
GraphLab's knowledge of the dependency structure allow modified
data to be communicated directly to the destination.



%
Overall, we can attribute GraphLab's superior performance over Hadoop
to the fact that the abstraction is a much better fit. It presents a simpler API to
the programmer and through the data graph, and informs GraphLab
about the communication needs of the program.

\textbf{MPI.} In order to analyze the cost of using a higher level
abstraction we implemented efficient versions of the Netflix and
NER applications using MPI.
The implementations made use of synchronous MPI collective operations
for communication.
The final performance results can be seen in
\figref{fig:alshadoop} and \figref{fig:coemhadoop}. We observe that the
performance of MPI and GraphLab implementations are
similar and conclude that GraphLab does not impose significant performance penalty.
GraphLab also has the advantage of being a higher level abstraction and is easier
to work with.

%
%
%
 
\tightsubsection{Locking Engine Evaluation} 

The CoSeg application makes use of dynamic prioritized scheduling
which requires the locking engine (\secref{sec:lockingengine}).  To
the best of our knowledge, there are no other abstractions which
provide the dynamic asynchronous scheduling as well as the
sequentially consistent sync (reduction) capabilities required by this
application.

\begin{figure*}[t]
        \centering
        \subfigure[CoSeg Weak Scaling] {
    \includegraphics[width=.23\textwidth]{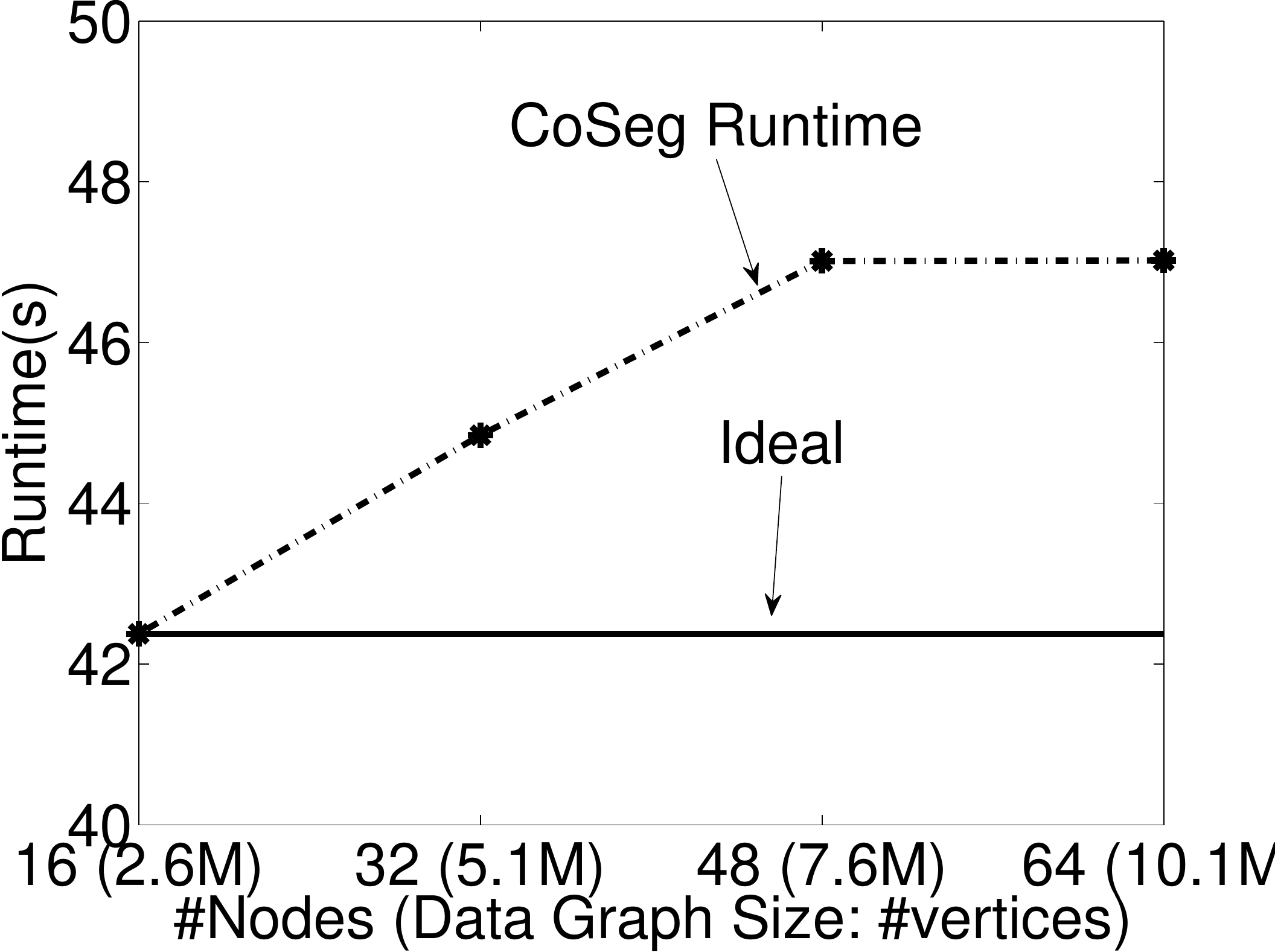}
  \label{fig:weakscalability}
    }
        \subfigure[Deferred Locking] {
        \raisebox{5mm}{
     \includegraphics[width=.23\textwidth]{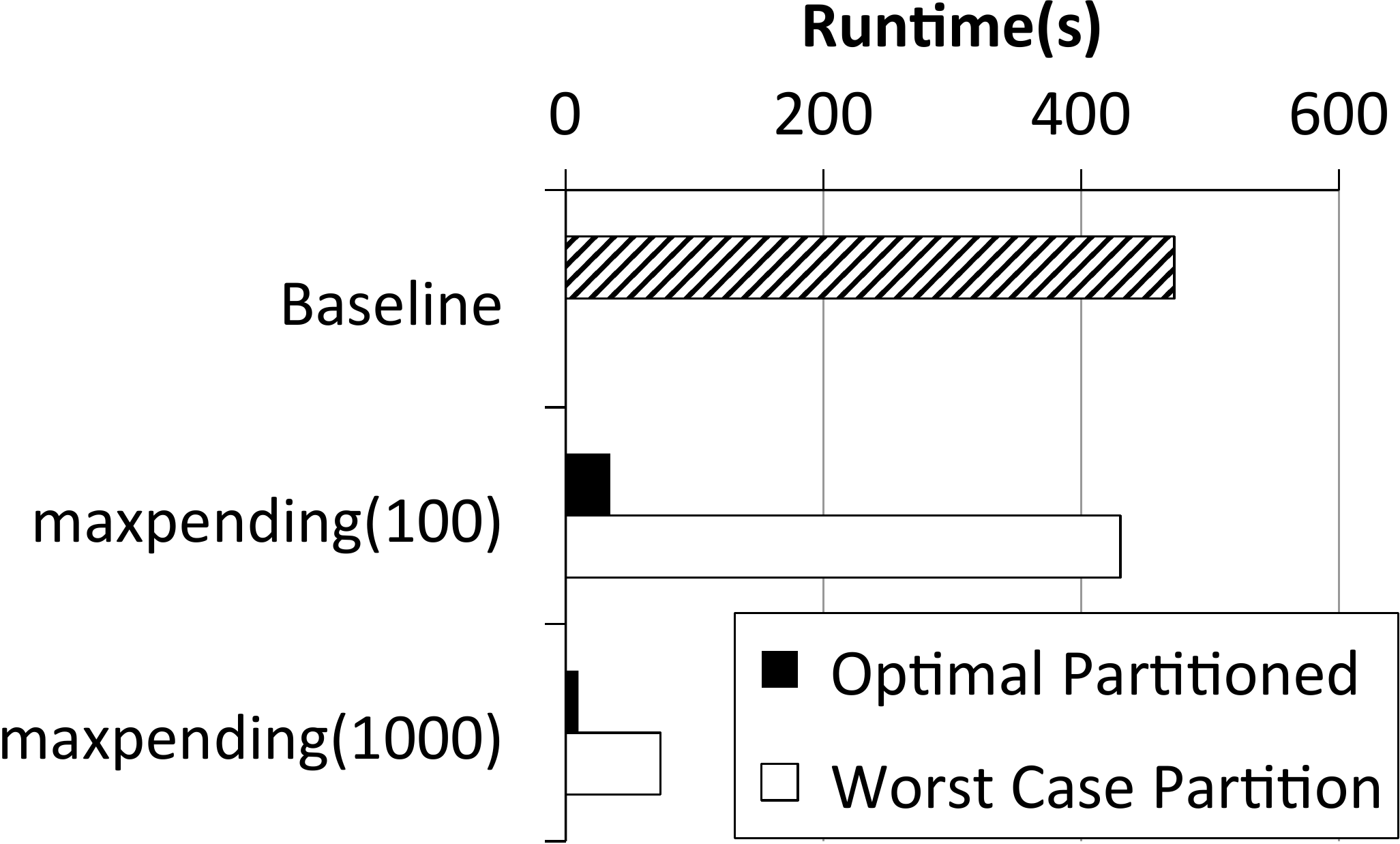}
     }
                \label{fig:lockevaluation}
    }
        \subfigure[EC2 Price/Performance]{
    \includegraphics[width=.23\textwidth]{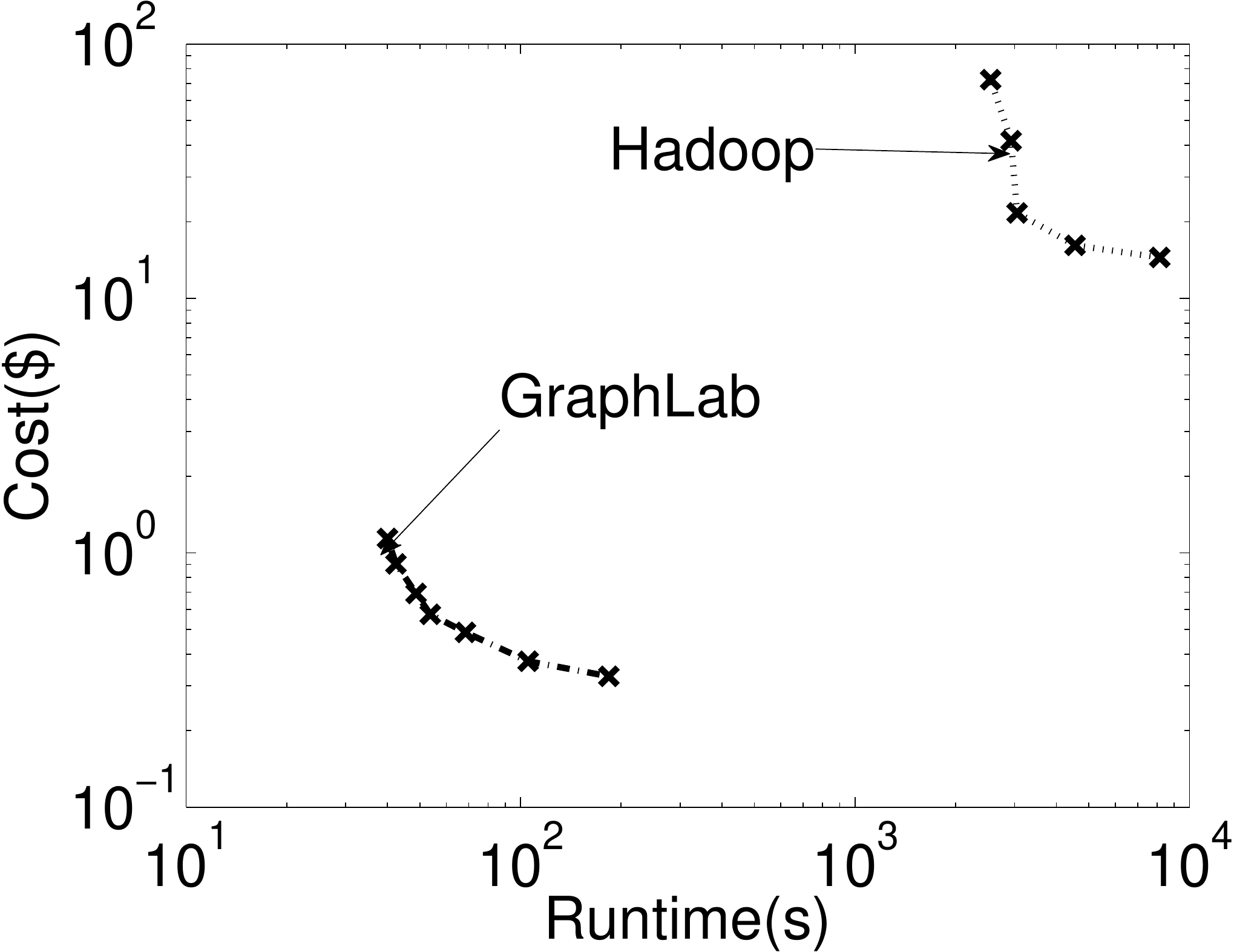}
  \label{fig:ec2cost}
}
        \subfigure[EC2 Price/Accuracy]{
    \includegraphics[width=.23\textwidth]{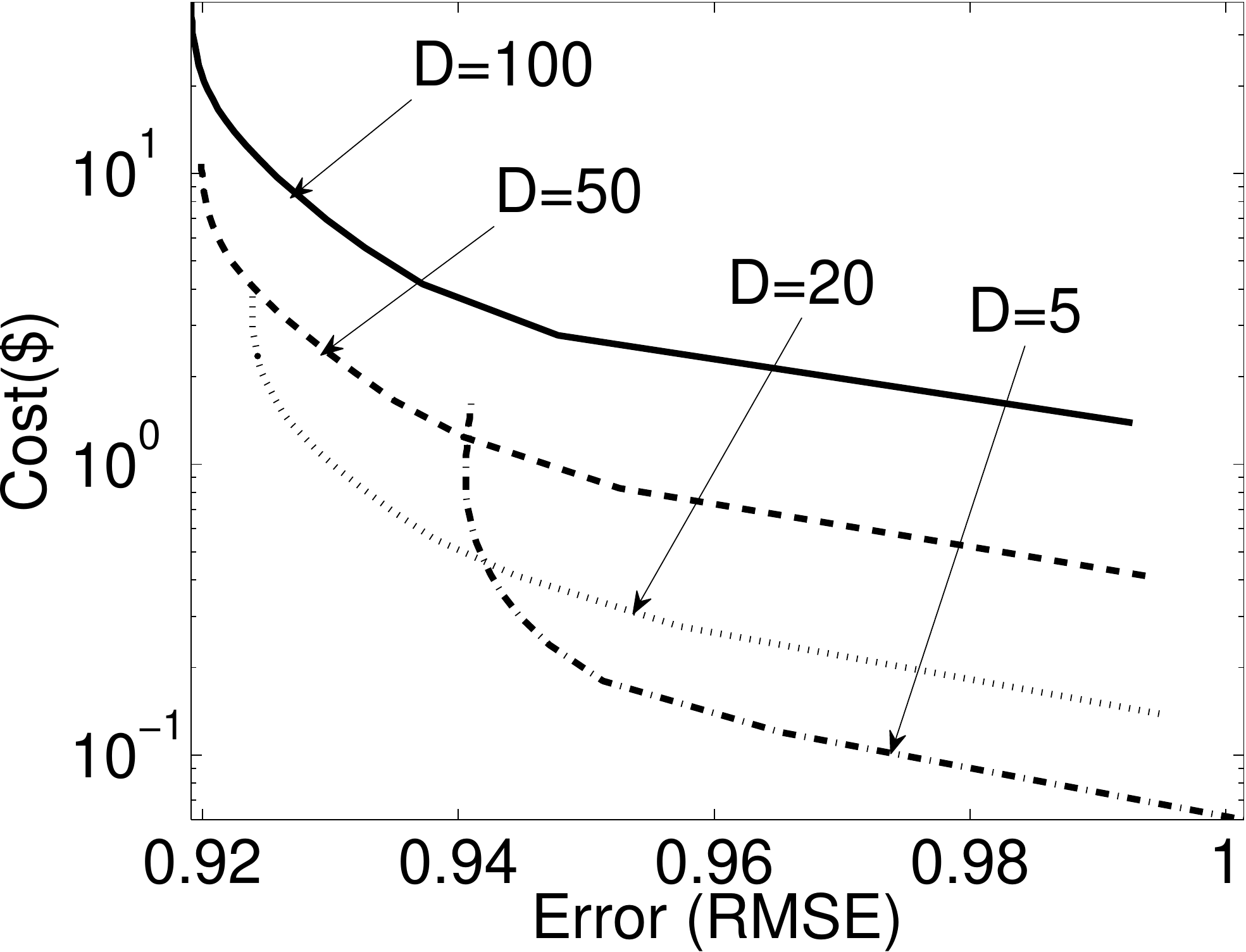}
  \label{fig:ec2accuracy}
    }
    \vspace{-3mm}
    \caption{\footnotesize 
      \textbf{(a)} Runtime of the CoSeg experiment as data set size is scaled together with
    the number of machines. Ideally the runtime should stay constant. GraphLab experiences only
    a small 11\% increase in runtime scaling up to 64 processors. 
     \textbf{(b)} The performance effects of varying the maximum
    number of pending locks (\texttt{maxpending}).
    When partitioning is good, increasing the number of pending locks has a small
    effect on performance. When partitioning is poor, increasing \texttt{maxpending} 
    improves performance significantly.
    \textbf{(c)} Price Performance ratio of GraphLab and Hadoop on Amazon EC2
    HPC nodes. Costs assume fine-grained billing. Note the log-log scale. Both Hadoop and GraphLab
    experience diminishing returns, but GraphLab is more cost effective.
    \textbf{(d)} Price Accuracy ratio of the Netflix experiment on
    HPC nodes. Costs assume fine-grained billing. Note the logarithmic cost scale. 
    Lower Error is preferred. 
}
    \vspace{-3mm}
\end{figure*}

In \figref{fig:speedupmain} we demonstrate that the locking engine can
provide significant scalability and performance on the large 10.5
million vertex graph used by this application, achieving a 10x speedup
with 16x more machines.
We also observe from \figref{fig:weakscalability} that the locking
engine provides nearly optimal weak scalability. The runtime does not
increase significantly as the size of the graph increases
proportionately with the number of processors. We can attributed this
to the properties of the graph partition where the number of edges
crossing machines increases linearly with the number of processors,
resulting in low communication volume.

In \figref{fig:lockevaluation} we further investigate the properties
of the distributed lock implementation described in
\secref{sec:lockingengine}. The evaluation is performed on a tiny
32-frame (192K vertices) problem on a 4 node cluster.  Two methods of
cutting the graph is explored. The first method is an ``optimal
partition" where the frames are distributed evenly in 8 frame blocks
to each machine. The second method is a ``worst case partition'' where
the frames are striped across the machines; this is designed to stress
the distributed lock implementation since every scope acquisition is
forced to acquire a remote lock. The maximum number of lock requests
allowed in the pipeline is varied (\texttt{maxpending}). The baseline
evaluation is optimal partitioning with the \texttt{maxpending} set to
zero.

%

\figref{fig:lockevaluation} demonstrates that on well-partitioned models, increasing the
maximum number of pending locks from $0$ to $100$ increases performance significantly. 
However, we observe diminishing returns as \texttt{maxpending} is further increased to $1000$.
On the other hand when the partitioning is poor, increasing the number of pending locks to $1000$ 
improves performance significantly.

%
%

\tightsubsection{EC2 Cost evaluation} To help put costs in
perspective, we plot a price-performance curve for the Netflix
application in \figref{fig:ec2cost}.  The curve shows the cost one
should expect to pay to obtain a certain desired performance level.
To ensure interpretability of the curve, the cost assumes fine-grained
billing even though Amazon EC2 billing rounds up utilization time to
the nearest hour.  The curve has an ``\textbf{L}'' shape implying
diminishing returns: as lower runtimes are desired, the cost of
attaining those runtimes increases faster than linearly. As a
comparison, we also provide the price-performance curve for Hadoop on
the same application. It is evident that for the Netflix application,
GraphLab is about two orders of magnitude more cost-effective than Hadoop.

\figref{fig:ec2accuracy} is an interesting plot which the cost
required to attain a certain degree of accuracy (lower RMSE is better)
on the Netflix task using 32 HPC nodes.  Similarly the curve
demonstrates diminishing returns: the cost of achieving lower test
errors increase quickly. The lower bound of all four curves inform the
reader with the ``cheapest" value of $d$ which attains the desired
accuracy.

\tightsection{Related Work}
\label{sec:relatedwork}
 
 Perhaps the closest approach to our abstraction is \textbf{Pregel}    \cite{pregel}, which also 
 computes on a user-defined data graph. The most important difference
 is that Pregel is based on the BSP model \cite{valiant1990bridging}, 
 while we propose an asynchronous model of computation. 
 Parallel BGL \cite{gregor2005parallel} is similar. 
 
\textbf{Piccolo} \cite{power2010piccolo} shares many similarities to GraphLab on an
implementation level, but does not explicitly model data dependencies. 
Sequential consistency of execution is also not guaranteed. 

\textbf {MapReduce} \cite{dean04} and \textbf{Dryad} \cite{isard2007dryad} are 
popular distributed data-flow frameworks which are used extensively in data mining. 
The use of MapReduce for ML in multi-core setting was first made popular by Chu et. al. \cite{cheng06} in 2006.
Such data-flow models cannot express efficiently sparse dependencies or iterative local computation. 
There are several extensions to these frameworks, such as \textbf{MapReduce Online} \cite{mapreduceonline}, 
\textbf{Twister} \cite{ekanayake2010twister}, \textbf{Nexus} \cite{hindman2009common} and \textbf{Spark} \cite{zaharia2010spark},
but none present a model which supports sparse dependencies with asynchronous local computation.
Most notably, \textbf{Surfer} \cite{chen2010large} extends MapReduce 
with a special primitive \emph{propagation} for edge-oriented tasks in graph processing,
but this primitive is still insufficient for asynchronous local computation.

Recently, work by \textbf{Pearce} et. al \cite{pearce2010multithreaded} 
proposed a system for asynchronous multithreaded graph traversals, including 
support for prioritized ordering. However, their work does 
does not address sequential consistency or the distributed setting. 
 
 Finally,   \textbf{OptiML} \cite{Chafi11}, a parallel programming language, 
  for Machine Learning. We share their approach of developing 
 domain specific parallel solutions.  OptiML parallelizes operations on linear algebra data structures, while 
 GraphLab defines a higher level model of parallel computation. 
 


\tightsection{Conclusion and Future Work}
\label{sec:conclusion}
Many important ML techniques utilize sparse computational 
dependencies, are iterative, and benefit from asynchronous computation.  
Furthermore, sequential consistency is an important requirement which
ensures statistical correctness and guarantees convergence for many 
ML algorithms. Finally, prioritized ordering of computation can 
greatly accelerate performance.  

The GraphLab abstraction we proposed allows the user to explicitly
represent structured dependent computation and extracts the available
parallelism without sacrificing sequential consistency. Furthermore,
GraphLab's sync operation allows global information to be efficiently
aggregated even as an asynchronous iterative algorithm proceeds.
Since the graph representation of computation is a natural fit for
many ML problems, GraphLab simplifies the design, implementation, and
debugging of ML algorithms.


We developed a highly optimized C++ distributed implementation of
GraphLab and evaluated it on three state-of-the-art ML algorithms
using real data: collaborative filtering on the Netflix dataset, Named
Entity Recognition, and Video Cosegmentation.  The evaluation was
performed on Amazon EC2 using up to 512 processors in 64 HPC nodes.
We demonstrated that GraphLab outperforms Hadoop (a popular framework
in the ML community) by 20-60x, and is competitive with tailored MPI
implementations.


Future work includes supporting dynamic and implicitly represented
graphs, as well as support for graphs in external storage.  The
current implementation provides limited support for external storage
through the use of \term{mmap}ed memory for vertex and edge data.
There are interesting possibilities for the intelligent placement and
caching of graph data to maximize performance
\cite{pearce2010multithreaded}.

While the current GraphLab implementation does not provide fault
tolerance, relatively simple modifications could be made to support
snapshotting capabilities. In particular, a globally consistent
snapshot mechanism can be easily performed using the Sync operation.
Additionally, we plan to extend GraphLab to other 
architectures including GPUs and supercomputers.


\vspace{-3mm}
{\denselistbib
  \footnotesize
 \bibliographystyle{unsrt}
  \bibliography{glsosp} 
}

\end{document}